\documentclass{article}

\usepackage[final, nonatbib]{neurips_2021}
\usepackage[nonatbib]{neurips_2021}

\usepackage[utf8]{inputenc} 
\usepackage{hyperref}       
\usepackage{url}            
\usepackage{booktabs}       
\usepackage{amsfonts}       
\usepackage{nicefrac}       
\usepackage{microtype}      
\usepackage{graphicx}
\usepackage{amssymb}
\usepackage{wrapfig}
\usepackage{thmtools}
\usepackage{thm-restate}
\usepackage{mathrsfs}
\usepackage{subcaption}
\usepackage{amsmath}
\usepackage{caption}
\usepackage{stmaryrd}
\usepackage{mwe}
\usepackage{mathtools}
\usepackage{colortbl}
\usepackage{wrapfig}
\usepackage{algorithm}
\usepackage[export]{adjustbox}
\usepackage{algorithmic}
\usepackage{amsthm}

\makeatletter
\newtheorem*{rep@theorem}{\rep@title}
\newcommand{\newreptheorem}[2]{%
\newenvironment{rep#1}[1]{%
 \def\rep@title{#2 \ref{##1}}%
 \begin{rep@theorem}}%
 {\end{rep@theorem}}}
\makeatother

\newtheorem{theorem}{Theorem}[section]
\newreptheorem{theorem}{Theorem}
\newtheorem{Lemma}{Lemma}[section]
\newreptheorem{lemma}{Lemma}
\newtheorem{prop}{Proposition}[section]
\newreptheorem{prop}{Proposition}
\newtheorem{defin}{Definition}[section]
\newreptheorem{defin}{Definition}

\title{DNN-based Topology Optimisation:  \\ Spatial Invariance and Neural Tangent Kernel}

\author{%
  Benjamin Dupuis \\
  Chair of Statistical Field Theory\\
  Ecole Polytechnique Fédérale de Lausanne\\
  Lausanne, Switzerland \\
  \texttt{benjamin.dupuis@epfl.ch} \\
  \And
  Arthur Jacot \\
  Chair of Statistical Field Theory\\
  Ecole Polytechnique Fédérale de Lausanne\\
  Lausanne, Switzerland \\
  \texttt{arthur.jacot@epfl.ch} \\
}

\begin{document}

\maketitle

\begin{abstract}
    We study the Solid Isotropic Material Penalisation (SIMP) method with a density field generated by a fully-connected neural network, taking the coordinates as inputs. In the large width limit, we show that the use of DNNs leads to a filtering effect similar to traditional filtering techniques for SIMP, with a filter described by the Neural Tangent Kernel (NTK). This filter is however not invariant under translation, leading to visual artifacts and non-optimal shapes. We propose two embeddings of the input coordinates, which lead  to (approximate) spatial invariance of the NTK and of the filter. We empirically confirm our theoretical observations and study how the filter size is affected by the architecture of the network. Our solution can easily be applied to any other coordinates-based generation method. 
\end{abstract}

\section{Introduction}

Topology optimisation \cite{topology}, also known as structural optimisation, is a method to find optimal shapes subject to some constraints. It has been widely studied in the field of computational mechanics. Here we are interested in the particular case of the Solid Isotropic Material Penalisation (SIMP) method \cite{topopt3D, topopt88}, which is a very common method in this field.

Recently some authors have used Deep Neural Networks (DNNs) to perform topology optimisation. We can differentiate two different approaches in the use of DNNs with SIMP. The first approach consists in generating with the classical algorithms a dataset of optimised shapes and train a DNN on this dataset to produce new optimal shapes \cite{CNN_topoptim, neural_topoptim}. Variations of this approach use Generative Adversarial Networks (GAN) \cite{topologyGAN, CGAN_topoptim} to effectively reproduce classical topology optimisation.

In the second approach, the density is generated pointwise by a DNN, which is trained with gradient descent to optimise the density field with respect to the physical constraints, as proposed in \cite{neural_param} to use the power of deep models without giving up exact physics. We focus on the approach of \cite{TOuNN, Fourier_TOUNN} where the density field is generated by a Fully-Connected Neural Network (FCNN) taking the coordinates of a grid as inputs. Surprisingly, \cite{TOuNN} observes that the DNN-generated density fields do not feature checkerboard artifacts, which are common in vanilla SIMP. A traditional method to avoid checkerboard patterns is to add a filter \cite{filtering, Bendsoe}, but it is not needed for DNN-generated density fields.


In this paper, we analyse theoretically how the use of a DNN to generate the density field affects the learning. Our main theoretical tool is the Neural Tangent kernel (NTK) introduced in \cite{NTK} to describe the dynamics of wide neural networks \cite{NTK, Arora, wide, hierarchy}.

While this paper focuses on linear elasticity and SIMP, our analysis can extended to other physical problems such as heat transfer \cite{heat}, or any model where an image is generated by a DNN taking the pixel coordinates as inputs (like in \cite{differentiable_image_param}).

\subsection{Our contribution}
In this paper we study topology optimisation with neural networks. The physical density is represented by a neural network taking an embedding of spatial coordinates as inputs, i.e. the density at a point $x \in \mathbb{R}^d$ is given by $f_{\theta}(\varphi(x))$ for $\theta$ the parameters of the network and $\varphi$ an embedding. We use theoretical tools, in particular the Neural Tangent Kernel (NTK), to understand how the architecture and hyperparameters of the network affect the optimisation of the density field:
\begin{itemize}

    \item We show that in the infinite width limit (when the number of neurons in the hidden layers grows to infinity), topology optimisation with a DNN is equivalent to topology optimisation with a density filter equal to the ``square root'' of the NTK. Filtering is a commonly used technique in topology optimisation, aimed to remove checkerboard patterns. 
    
    \item In topology optimisation as in other physical optimisation problems, it is crucial to guarantee some spatial invariance properties. If the coordinates are taken as inputs of the network directly, the NTK (and the corresponding filter) is not translation invariant, leading to non-optimal shapes and visual artifacts. We present two methods to ensure the spatial invariance of the NTK: embedding the coordinates on the (hyper-)torus or using a random Fourier features embedding (similar to \cite{Fourier_features}).
    
    \item In traditional topology optimisation, the filter size must be tuned carefully. When optimising with a DNN, the filter size depends on the embedding of the coordinates and the architecture of the network. We define a filter radius for the NTK, which plays a similar role as the classical filter size and discuss how it is affected by the choice of embedding, activation function, depth and other hyperparameters like the importance of bias in the network.
    This tradeoff can also be analysed in terms of the  spectrum of the NTK, explaining why neural networks naturally avoid checkerboard patterns.
    
\end{itemize}

We confirm and illustrate these theoretical observations with numerical experiments. Our implementation of the algorithm will be made public at \url{https://github.com/benjiDupuis/DeepTopo}.

\section{Presentation of the method}
In this paper, we use a DNN to generate the density field used by the Solid Isotropic Material Penalisation (SIMP) method. Our implementation of SIMP is based on \cite{topopt88} and \cite{topopt3D}. In this section we introduce the traditional SIMP method and our neural network setting.

\subsection{SIMP method}
 We consider a regular grid of $N$ elements where the density of element $i$ is denoted $y_i \in [0,1]$, informally the value $y_i$ represents the presence of material at a point $i$. Our goal is to optimise over the density $y \in \mathbb{R}^N$ to obtain a shape that can withstand forces applied at certain points, represented by a vector $F$.

The method uses finite element analysis to define a stiffness matrix $K(y) \in S_N^{++}(\mathbb{R})$ from the density $y$ and computes the displacement vector $U(y)$ (which represent the deformation of the shape at all points $i$ as a result of the applied forces $F$) by solving a linear system $K(y)U(y) = F$. In our implementation, we performed it either by using sparse Cholesky factorisation \cite{cholmod, cholmod2} or BICGSTAB method \cite{bicgstab} (this last one can be used for a high number of pixels). 

The loss function is then defined as the compliance $C(y) = U(y)^T K(y) U(y)$, under a volume constraint of the form $\sum_{i=1}^N y_i = V_0$, with $0 \leq V_0 \leq N$ (see \cite{topopt88, topopt3D}). 


\subsection{A modified SIMP approach}

Several methods exist to optimise the density field $y \in \mathbb{R}^N$, such as gradient descent or the so-called Optimality Criteria (OC) \cite{optimality_criteria}. We propose here an optimisation method inspired from \cite{neural_param} which we will refer as the Modified Filtering method (MF). The advantage of this method is that it can be used with or without DNNs, hence allowing comparison between these two approaches. We first present here the model without DNNs. 

In our method, the densities $y_i^{\text{MF}}$ are given by:
\begin{equation}
\forall i \in \{1,...,N\}, ~y_i^{\text{MF}} = \sigma(x_i + \bar{b}(X)), \quad \text{with }\bar{b}(X)\text{ such that } \sum_{i=1}^N y_i^{\text{MF}} = V_0,
\label{optimal_bias}
\end{equation}
for $X = (x_1,...,x_N) \in \mathbb{R}^N$ and the sigmoid $\sigma(x) = \frac{1}{1 + e^{-x}}$. We will denote this operation as: $Y^{\text{MF}} = \Sigma(X)$. The sigmoid ensures that densities are in $[0,1]$ and the choice of the optimal bias $\bar{b}(X)$ ensures that the volume constraint is satisfied.

\textbf{Filtering:} If the vector $X$ is optimised directly with gradient descent, SIMP often converges toward checkerboard patterns, i.e. some high frequency noise in the image, which is a common issue with SIMP \cite{topopt88}. To overcome this issue a common technique is to use filtering \cite{filtering}. In this paper, we consider low-pass density filters of the form: $X = T\bar{X}$ where $T$ represents a convolution on the grid, $\bar{X}$ are the design variables and $X$ is the vector in equation \ref{optimal_bias}.
The loss function of this method is then naturally defined as:
$
\bar{X} \longmapsto C(\Sigma(T\bar{X})).
$

 The gradient $\nabla_Y C$ is easily obtained by the self-adjointness of the variational problem \cite{optimality_criteria, metasurfaces}. We  recover $\nabla_X C$ from $\nabla_Y C$ using an implicit differentiation technique \cite{implicit_diff}. The following proposition is a consequence of implicit function theorem and chain rules:
\begin{prop}
\label{implicit_diff}
    Let $\dot{S}$ be the vector with entries $\dot{\sigma}(x_i + \bar{b}(X))$. We have $\nabla_X C = D_X \nabla_Y C$ with:
    \begin{equation}
    D_X := -\frac{1}{\vert \dot{S} \vert_1}\dot{S}\dot{S}^T + \text{Diag}(\dot{S}).
    \end{equation}
    where $\vert . \vert_1$ denotes the $l^1$ norm of a vector. Furthermore $D_X$ is a symmetric positive semi-definite matrix whose null-space is the space of constant vectors and has eigenvalues smaller than $\frac{1}{4}$.
\end{prop}

\subsection{Proposed algorithm: SIMP with Neural networks}
\label{SIMP_with_DNNs}
 Fully-Connected Neural Networks (FCNN) are characterised by the number of layers $L+1$, the numbers of neurons in each layer $(n_0,n_1,..., n_L)$ and an activation function $\mu : \mathbb{R} \longrightarrow \mathbb{R}$, here we will use the particular case $n_L=1$. The activations $a^l \in \mathbb{R}^{n_l}$ and preactivations $\Tilde{a}^l \in \mathbb{R}^{n_l}$ are defined recursively for all layers $l$, using the so-called NTK parameterisation \cite{NTK}:
\begin{equation}
    \label{DNN}
a^{0}(x) = x, \quad \Tilde{a}^{l+1}(x) = \frac{\alpha}{\sqrt{n_l}} W^l a^l(x) + \beta b^l, \quad a^{l+1}(x) = \mu \big( \Tilde{a}^{l+1}(x) \big),
\end{equation}
for some hyperparameters $\alpha,\beta \in [0,1]$ representing the contribution of the weights and bias terms respectively. The parameters $\theta = (\theta_p)_p$, consisting in weight matrices $W^l$ and bias vectors $b^l$ are drawn as i.i.d. standard normal random variables $\mathcal{N}(0,1)$. We denote the output of the network as $f_{\theta}(x) = \Tilde{a}^L(x)$.

\textbf{Remark:} To ensure that the variance of the neurons at initialization is the equal to 1 at all layers, we choose $\alpha$ and $\beta$ such that $\alpha^2 + \beta^2 = 1$ and use a standardised non-linearity, i.e. $\mathbb{E}_{X \sim \mathcal{N}(0,1)}[\mu(X)^2] = 1$ (\cite{order_chao}).


\begin{figure}[!h]
    \centering
    \includegraphics[trim={0.5cm 4cm 0cm 4cm},clip, scale=0.57]{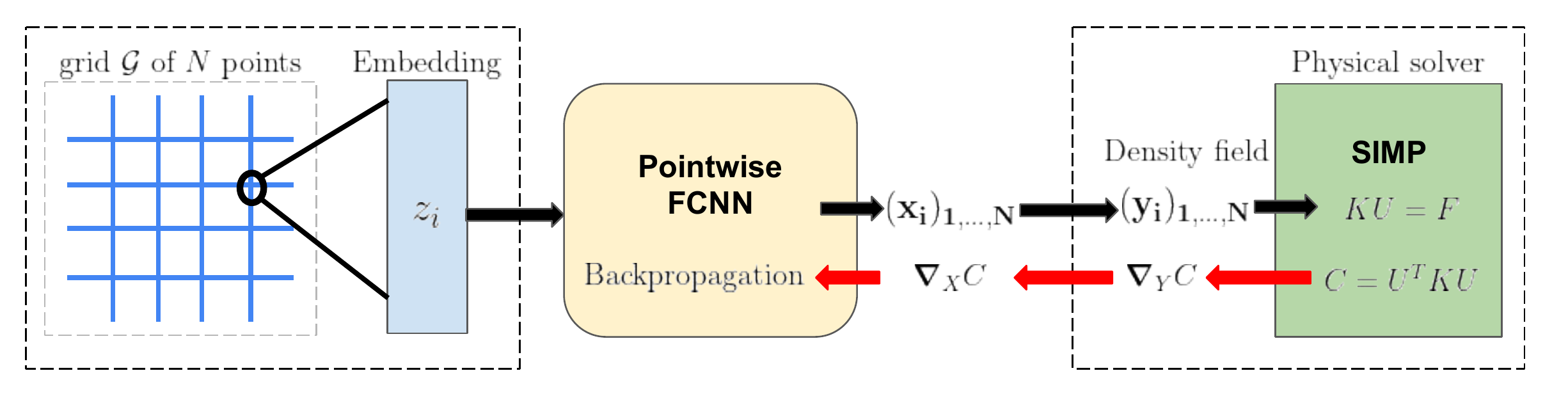}
    \caption{Illustration of our method}
    \label{schema}
\end{figure}

In our approach, the pre-densities $X^{\text{NN}}(\theta) = (x_1^{\text{NN}},...,x_N^{\text{NN}})$ are generated by a neural network as $ x_i^{\text{NN}} = f_\theta (z_i)$
where $z_i \in \mathbb{R}^{n_0}$ is either the coordinates of the grid elements (in this case $n_0=d$) or an embedding of those coordinates. We then apply the same transformation $\Sigma$ to obtain the density field $Y^{\text{NN}}(\theta) = \Sigma(X^{\text{NN}}(\theta))$. Our loss function is then defined as:
$$
\theta \longmapsto C(Y^{\text{NN}}(\theta)) = C \big( \Sigma(X(\theta)) \big).
$$
The design variables are now the parameters $\theta$ of the network. The gradient $\nabla_\theta C$ w.r.t. to the parameters is computed by first using Proposition \ref{implicit_diff} to get $\nabla_{Y^{\text{NN}}}C$ followed by traditional backpropagation.

\textbf{Remark:} Note the absence of filter $T$ in the above equations, indeed we will show how neural networks naturally avoid checkerboard patterns, making the use of filtering obsolete.

\textbf{Initial density field:} The SIMP method is usually initialised with a constant density field \cite{topopt88}. Since the neural network is initialized randomly, the initial density field is random and non-constant. To avoid this problem, we subtract the initial density field and add a well-chosen constant:
\begin{equation}
\label{initial_volume}
\forall i \in \{1,...,N\},~x_i(\theta) = \bar{f}_{\theta(t)}(z_i) = f_{\theta(t)}(z_i) - f_{\theta(t=0)}(z_i) + \log\bigg( \frac{V_0}{N - V_0}\bigg).
\end{equation}
We used equation \ref{initial_volume} to compute $X(\theta)$ in our numerical experiments.



\begin{figure}[!h]
    \centering
    \includegraphics[width = 1.\textwidth, clip, trim=5cm 0cm 3cm 0cm]{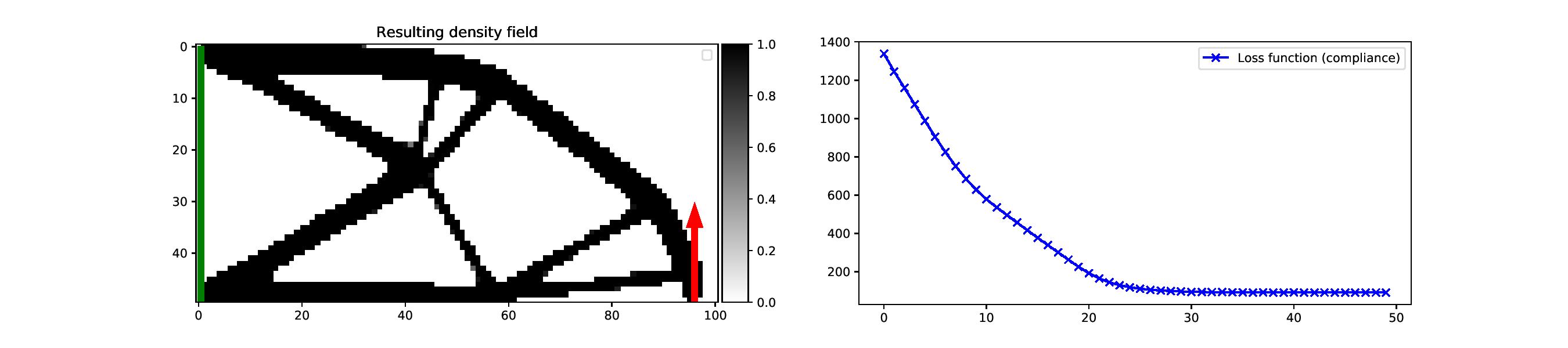}
    \caption{Example of result of our method with applied forces (red arrow) and a fixed boundary (green). Here we used a Gaussian embedding (see section $4$ for details).}
    \label{result}
\end{figure}


\section{Theoretical Analysis}

\subsection{Analogy between the Neural Tangent Kernel and filtering techniques}

In our paper, we use the Neural Tangent Kernel (NTK \cite{NTK}) as the main tool to analyse the training behaviour of the FCNN. In our setting (where $n_L = 1$) the NTK is defined as:
$$
\forall z,z' \in \mathbb{R}^{n_0},~\Theta^L_{\theta}(z,z') = \sum_{p} \frac{\partial f_{\theta}}{\partial \theta_p}(z) \frac{\partial f_{\theta}}{\partial \theta_p}(z') = (\nabla_{\theta}f_{\theta}(z) \vert \nabla_{\theta}f_{\theta}(z')).
$$
This is a positive semi-definite kernel. Given some inputs $z_1,...,z_N$ we define the NTK Gram matrix as:
$
\Tilde{\Theta}^L_{\theta} := \big( \Theta^L(z_i,z_j) \big)_{1 \leq i,j \leq N} \in \mathbb{R}^{N \times N}.
$


Assuming a small enough learning rate, the evolution of the network under gradient descent is well approximated by the gradient flow dynamics $\partial_t \theta(t) = -\nabla_{\theta}C(t)$. The evolution of the output of the network $X^{\text{NN}}(\theta)$ can then easily be expressed in terms of the NTK Gram matrix \cite{order_chao} for a loss $\mathcal{L}$:
$$
\partial_t X^{\text{NN}}(\theta(t)) = - \Tilde{\Theta}^L_{\theta (t)} \nabla_{X^{\text{NN}}} \mathcal{L} .
$$
From this equation we can derive the evolution of the physical density field $Y^{\text{NN}}$ in our algorithm:
\begin{prop}
    \label{evolution_eq}
    If the network is trained under this gradient flow, then by applying chain rules, we can prove that the density field follows the equation:
    \begin{equation}
       \partial_t Y^{\text{NN}}(\theta(t)) = - D_{X}(t) \Tilde{\Theta}^L_{\theta (t)} D_{X}(t) \nabla_{Y} C(Y^{\text{NN}}(\theta(t))).
    \end{equation}
\end{prop}

The analogy between the NTK and filtering techniques comes from the following observation. With Modified Filtering with a filter $T$, we show similarly that the density field $Y^{\text{MF}}$ evolves as
\begin{equation}
    \label{evolution_filter}
    \partial_t Y^{\text{MF}} (t) = -D_X(t) T T^T D_X(t) \nabla_Y C(Y^{\text{MF}}(t)) .
\end{equation}

We see that the NTK Gram matrix and the squared filter $TT^T$ play exactly the same role. An important difference however is that the NTK is random at initialisation and evolves during training.

This difference disappears for large widths (when $n_1,\dots ,n_{L-1}$ are large), since the NTK converges to a deterministic and time independent limit $\Tilde{\Theta}_{\infty}^L$ as $n_1,\dots ,n_{L-1}\to\infty$ \cite{NTK}. Furthermore, in contrast to the finite width NTK (also called empirical NTK), we have access to a closed form formula for the limiting NTK $\Tilde{\Theta}_{\infty}^L$ (given in the appendix).

In the infinite width limit, the evolution of the physical densities is then expressed in terms of the limiting NTK Gram matrix $\Tilde{\Theta}_{\infty}^L$:
\begin{equation}
   \partial_tY^{\text{NN}}(\theta(t)) = - D_{X}(t) \Tilde{\Theta}_{\infty}^L D_{X}(t) \nabla_{Y} C(Y^{\text{NN}}(\theta(t))).
    \label{evolution_limit}
\end{equation}
From now on we will focus on this infinite-width limit,  comparing the NTK Gram matrix $\Tilde{\Theta}_{\infty}^L$ and the squared filter $TT^T$. Recent results \cite{wide, Arora, hierarchy} suggest that this limit is a good approximation when the width of the network is sufficiently large. For more details see the appendix, where we compare the empirical NTK with its limiting one and plot its evolution in our setting.


\subsection{Spatial invariance}
\label{spatial_invariance}
In physical problems such as topology optimisation, it is important to ensure that certain physical properties are respected by the model. We focus in this section on the translation and rotation invariance of topology optimisation: if the force constraints are rotated or translated, the resulting shape should remain the same (up to rotation and translation), as in Figure \ref{big} (b.1 and b.2).

In Modified Filtering method, this property is guaranteed if the filter $T$ is translation and rotation invariant. In contrast the limiting NTK is in general invariant under rotation \cite{NTK} but not translation. As Figure \ref{big} shows, this leads to some problematic artifacts. The NTK can be made translation and rotation invariant by first applying an embedding $\varphi : \mathbb{R}^d \longrightarrow \mathbb{R}^{n_0}$ with the properties that for any two coordinates $p,p'$, $\varphi(p)^T\varphi(p')$ only depends on the distance $\Vert p - p' \Vert_2$. Since the rotation invariance of the NTK implies that $\Theta^L_\infty(z,z')$ depends only on the scalar products $z^Tz'$, $zz^T$ and $z'z'^T$, we have that $\Theta^L_\infty(\varphi(p),\varphi(p'))$ depends only on  $\Vert p - p' \Vert$ as needed.


The issue is that for finite $n_0$ there is no non-trivial embedding $\varphi$ with this property:
\begin{prop}
\label{embedding}
Let $\varphi:\mathbb{R}^d \to \mathbb{R}^{n_0}$ for $d>2$ and any finite $n_0$. If $\varphi$ satisfies $\varphi(x)^T\varphi(x')=K( \Vert x-x' \Vert)$ for some continuous function $K$ then both $\varphi$ and $K$ are constant.
\end{prop}

To overcome this issue, we present two approaches to approximate spatial invariance with finite embeddings: an embedding on a (hyper)-torus and a random feature \cite{random_features} embedding based on Bochner theorem \cite{bochner}.

\subsubsection{Embedding on a hypertorus}
In this subsection we consider the following embedding of a $n_x \times n_y$ regular grid on a torus:
\begin{equation}
    \label{torus_embedding_formula}
    \mathbb{R}^2 \ni p  = (p_1, p_2) \longmapsto \varphi(p) = r(\cos(\delta p_1), \sin(\delta p_1), \cos(\delta p_2), \sin(\delta p_2)),
\end{equation}
where $\delta > 0$ is a discretisation angle (our default choice is $\delta = \frac{\pi}{2 \max(n_x, n_y)}$). One can use similar formulas for $d > 2$ (leading to an hyper-torus embedding), we used $d=2$ in equation \ref{torus_embedding_formula} for simplicity.

This embedding leads to an exact translation invariance and an approximate rotation invariance:
$$
\varphi(p)^T\varphi(p') = r^2 (\cos(\delta(p_1 - p_1')) + \cos(\delta(p_2 - p'_2))) = r^2 \bigg( 2 - \frac{\delta^2}{2}\Vert p - p' \Vert_2^2 \bigg) + \mathcal{O}\big( \delta^4  \Vert p - p' \Vert_4^4 \big).
$$

As a result, the limiting NTK $\Theta_{\infty}(\varphi(p), \varphi(p'))$ is translation invariant and approximately rotation invariant (for small $\delta$ and/or when $p,p'$ are close to each other). Moreover, if we look at the limiting NTK on the whole torus, we obtain that the gram matrix $\Tilde{\Theta}_{\infty}$ is a discrete convolution on the input grid, with nice properties summed up in the following proposition:
\vspace{2.5pt}

\begin{prop}
    \label{torus_prop}
    We can always extend our $n_x \times n_y$ grid and choose $\delta$ such that the embedded grid covers the whole torus (typically $\delta = \frac{\pi}{2 \max(n_x, n_y)}$ and take a $n \times n$ grid with $n = 4\max(n_x, n_y)$). Then the Gram matrix $\Tilde{\Theta}_{\infty}$ of the limiting NTK is a $2$D discrete convolution matrix. Moreover the NTK Gram matrix has a positive definite square root $\sqrt{\Tilde{\Theta}_{\infty}}$ which is also a discrete convolution matrix.
\end{prop}

\begin{wrapfigure}[11]{r}{0.5\textwidth}
\centering
    \includegraphics[width=0.48\textwidth, trim= 1cm 1cm 1cm 2cm]{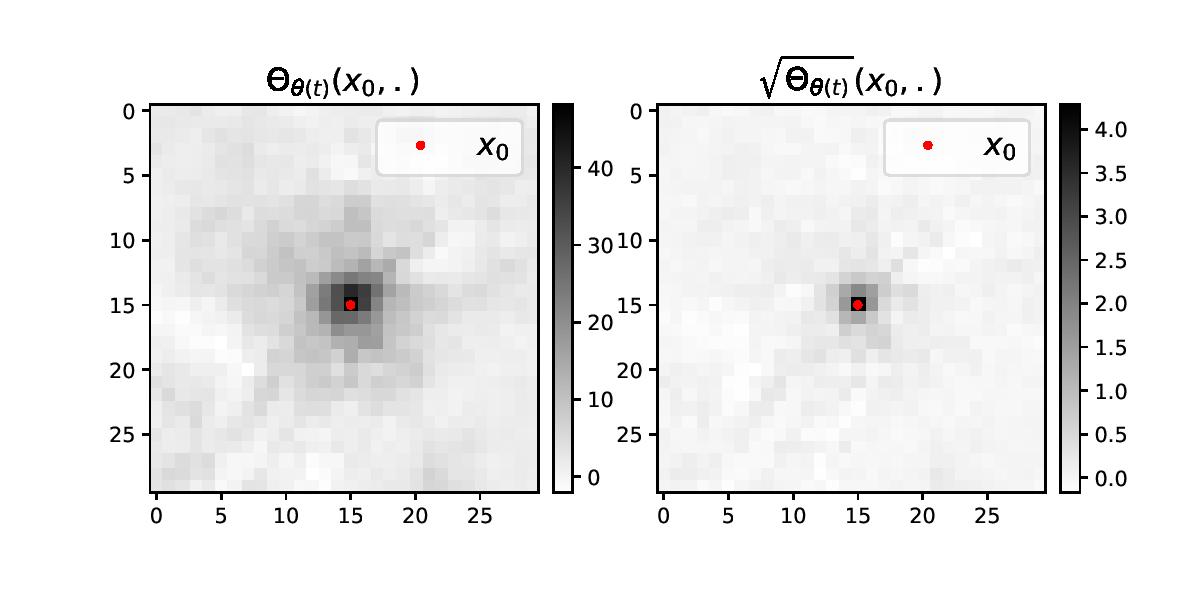}
  \caption{Representation of one line of $\Tilde{\Theta}_{\theta}$ on the full torus and of its square root. We used $\beta=0.2$ and $\omega = 3$ (see Section \ref{experimental_setup}) here to make the filter visible on the whole torus.}
  \label{square_root_full_torus}
\end{wrapfigure}

As we know, the eigenvectors of such a convolution matrix are the $2$D Fourier vectors. The corresponding eigenvalues are the discrete Fourier transforms of the convolution kernel.

The square root of the NTK Gram matrix $\sqrt{\Tilde{\Theta}_{\infty}}$ then corresponds to the filtering matrix $T$ in our analogy. Figure \ref{square_root_full_torus} shows that on the full torus, the matrix square root $\sqrt{\Tilde{\Theta}_{\theta}}$ indeed looks like a typical smoothing filter. 

\begin{figure}
    \centering
    \boxed{\includegraphics[width=0.98\linewidth, clip, trim=4cm 1cm 2cm 1cm]{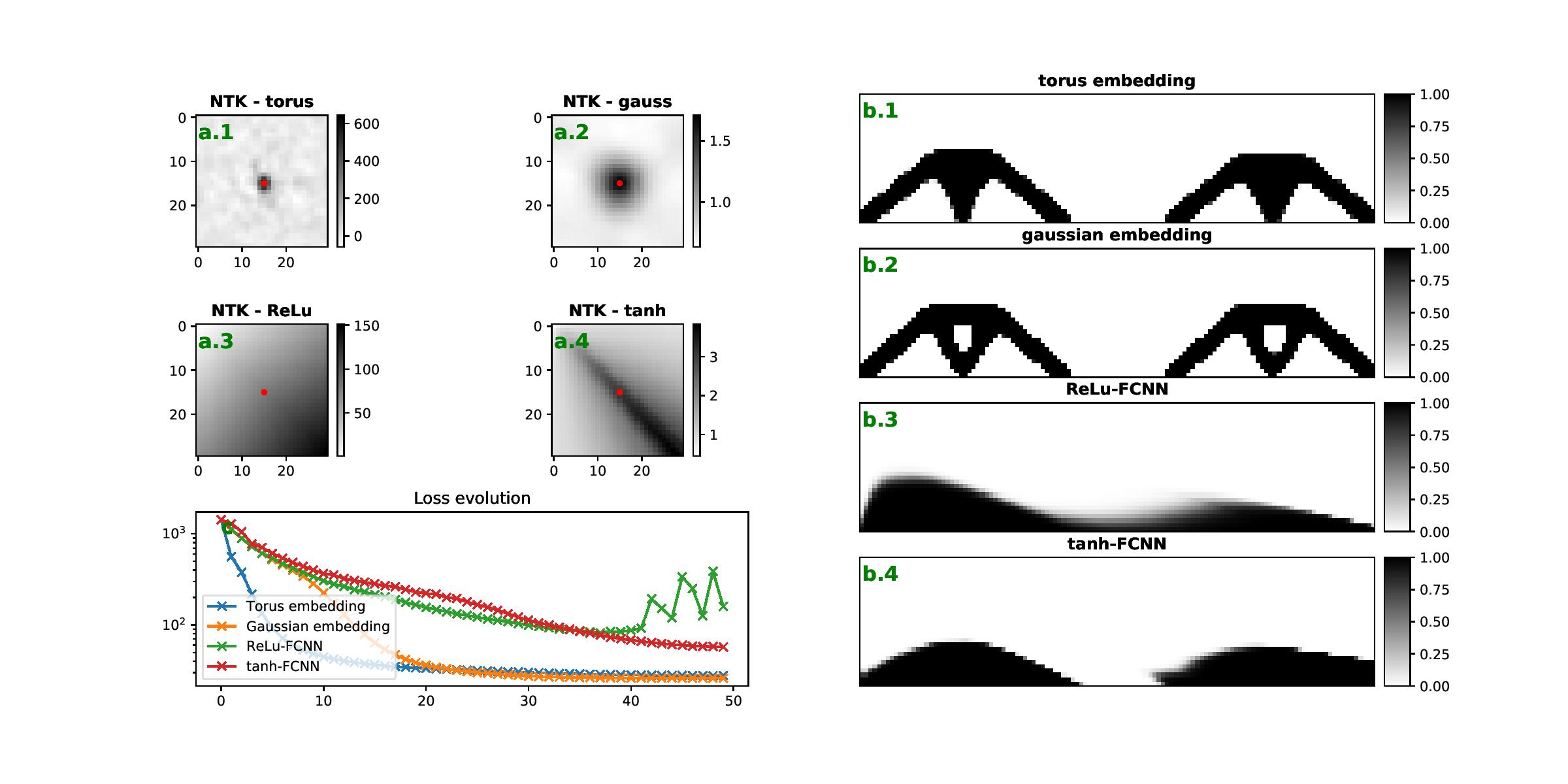}}
    \caption{Left: empirical NTK of FCNNs with both embedding (a.1, a.2, see Section \ref{experimental_setup} for details) or without embedding (a.3 with ReLu, a.4 with tanh). Right: Corresponding shape obtained after training. Note that methods without spatial invariance particularly struggles with this symmetric load case (b.3, b.4) while both "embedded methods" respect the symmetry (b.1, b.2). We also observed that training with non-embedded methods is very unstable}
    \label{big}
\end{figure}

As Figure \ref{big} shows, the torus embedding method gives good numerical results and respect the symmetry of the applied forces $F$.


\subsubsection{Random embeddings for radial kernels}
\label{random_embedding}
Another approach to approximate a rotation and translation invariant embedding is to use random Fourier features \cite{random_features}, which is a general method to approximate shift invariant kernels of the form $k(x,y) = k(x - y)$. By  Bochner theorem \cite{bochner}, any continuous non-zero radial kernel $k(x-y) = K(\Vert x - y \Vert)$ can be written as the the (scaled) Fourier transform of a probability measure $\mathbb{Q}$ on $\mathbb{R}^d$:
$$
k(r)= k(0) \int_{\mathbb{R}^d} e^{i \omega . r} d\mathbb{Q}(\omega).
$$
For radial kernels, we formulate random Fourier features embeddings $\varphi:\mathbb{R}^d\to\mathbb{R}^{n_0}$ as follows:
$$
\varphi(p)_i =\sqrt{2 k(0)} \sin(w_i^T p + \frac{\pi}{4}+  b_i),
$$
for i.i.d. samples $w_1,...,w_{n_0} \in \mathbb{R}^d$ from $\mathbb{Q}$ (which is also invariant by rotation) and i.i.d. samples $b_1,...,b_{n_0} \in \mathbb{R}^d$ from any symmetric probability distribution (or uniform laws on $[0,2\pi]$). By the law of large numbers for large $n_0$, we have the approximation $\frac{1}{n_0}\varphi(p)^T \varphi(p') \simeq k(p - p')$.

\textbf{Gaussian embedding:} Depending on the kernel $k$ that we want to approximate, it may be difficult to sample from the distribution $\mathbb{Q}$. The simplest case is for a Gaussian kernel $k(d)=e^{-\frac{1}{2\ell^2}d^2}$, where the distribution $\mathbb{Q}$ of the weights $w_i$ is $\mathcal{N}(0,\frac{1}{\ell^2}I_d)$, i.e. the entries $w_{ij}$ are all i.i.d. $\mathcal{N}(0,\frac{1}{\ell^2})$ Gaussians.
For this reason this is the embedding that we will use in our numerical experiments.
Note the similarity between this type of embedding and an untrained first layer of a FCNN with sine activation function, weights $w_i$ and bias $b_i$.

Moreover, the following result shows that  we can still define a "square root" of the NTK with those types of embedding and thus complete the analogy with equation \ref{evolution_filter}.

\begin{prop}
    \label{continuous_square_root}
    Let $\varphi$ be an embedding as described above for a positive radial kernel $k\in L^1(\mathbb{R}^{d})$ with $k(0) = 1$, $k \geq 0$. Then there is a filter function $g:\mathbb{R}\to\mathbb{R}$ and a constant $C$ such that for all $p,p'$:
    \begin{equation}
        \label{square_root}
    \lim_{n_0\to\infty}\Theta_{\infty}(\varphi(p),\varphi(p')) = C + (g \star g)(p-p'),
    \end{equation}
    where $\Theta_\infty$ is the limiting NTK of a network with a Lipschitz, non-constant and standardised activation function $\mu$. (Here $\star$ denotes the convolution product).
\end{prop}
As the matrix $D_X$ in equation \ref{evolution_limit} cancels out the constant frequency (proposition \ref{implicit_diff}), the constant $C$ doesn't matter, i.e. $D_X\tilde{\Theta}_\infty^{(L)}D_X = D_X\left(\tilde{\Theta}_\infty^{(L)} - C\right)D_X$.

\section{Experimental analysis}


\subsection{Setup}
\label{experimental_setup}
Most of our experiments were conducted with a torus embedding or a gaussian embedding. For the SIMP algorithm, we adapted the code described in \cite{topopt88, topopt3D}. Here are the hyperparameters used in the experiments. 

For the Gaussian embedding, we used $n_0=1000$ and a length scale $\ell = 4$. This embedding was followed by one hidden linear layer of size $1000$ with standardized ReLu ($x \mapsto \sqrt{2}\max(0,x)$)  and a bias parameter $\beta = 0.5$. 


For the torus embedding we set the torus radius to $r = \sqrt{2}$ (to be on a standard sphere) and the discretisation angle to $\delta = \frac{\pi}{2 \max(n_x ,n_y)}$ (to cover roughly half the torus, which is a good trade-off between rotation invariance and kernel size), where $n_x\times n_y$ is the size of the grid.  It was followed by $2$ linear layers of size $1000$ with $\beta = 0.1$. The ReLu activation is not well-suited in this case because it induces filters that are too wide. The large radius of the NTK kernel can be understood in relation with the order/chaos regimes \cite{Deep_Info_Prop_Schoenholz2017, Chaos_Poole2016}, as observed in \cite{order_chao}  the ReLU lies in the ordered regime when $\beta>0$, leading to a ``wide'' kernel, a narrower kernel can be achieved with non-linearities which lie in the chaotic regime instead. We used a cosine activation of the form $x \mapsto \cos(\omega x)$, which has the advantage that the width of the filter can be adjusted using the $\omega$ hyperparameter, see Section \ref{filter_radius}. When not stated otherwise we used $\omega=5$.

Even though our theoretical analysis is for gradient flow, we obtain similar results with other optimizers such as RPROP \cite{Rprop} (learning rate $10^{-3}$) and ADAM \cite{adam} (learning rate $10^{-3}$). RPROP gave the fastest results, possibly because it is well-suited for batch learning \cite{Rprop_full_batch}.  Vanilla gradient descent can be very slow due to the vanishing of the gradients when the image becomes almost binary (due to the sigmoid), we therefore gradually increased the learning rate during training to compensate. 




\subsection{Spectral analysis}

In SIMP convolution with a low pass filter ensures that low frequencies are optimised faster than high frequencies, to avoid checkerboards.

\begin{figure}[t]
\centering
\begin{minipage}[b]{0.48\textwidth}
   \centering
    \includegraphics[width=0.85\textwidth, clip, trim=2.cm 2cm 2cm 3cm]{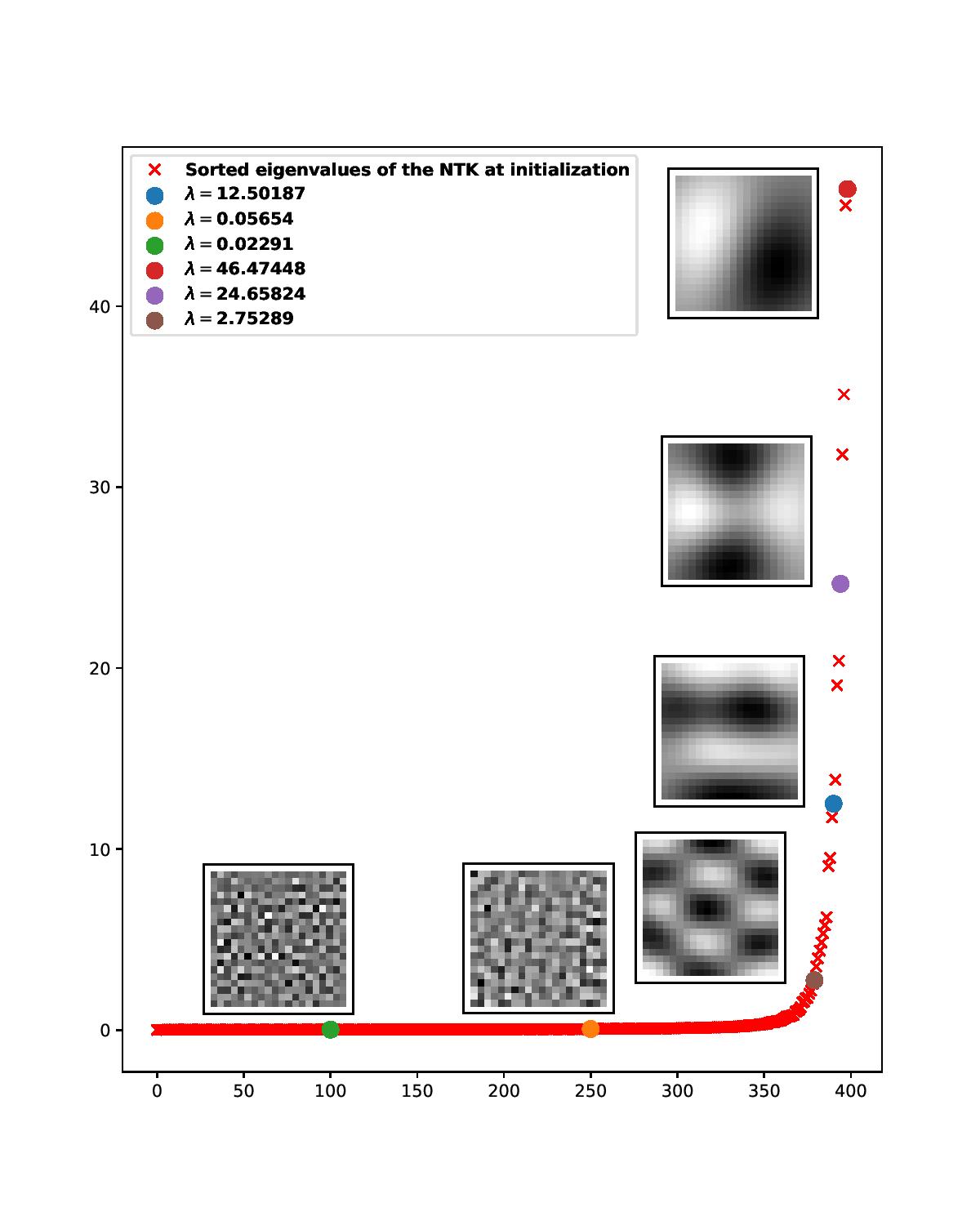}
  \caption{Sorted eigenvalues of the empirical NTK with some eigenvectors (reshaped as images). Obtained with a Gaussian embedding.}
  \label{spectral}
\end{minipage}
\hfill
\begin{minipage}[b]{0.48\textwidth}
    \centering
     \includegraphics[width=0.88\textwidth, trim=2.cm
     2.5cm 4cm 6.5cm]{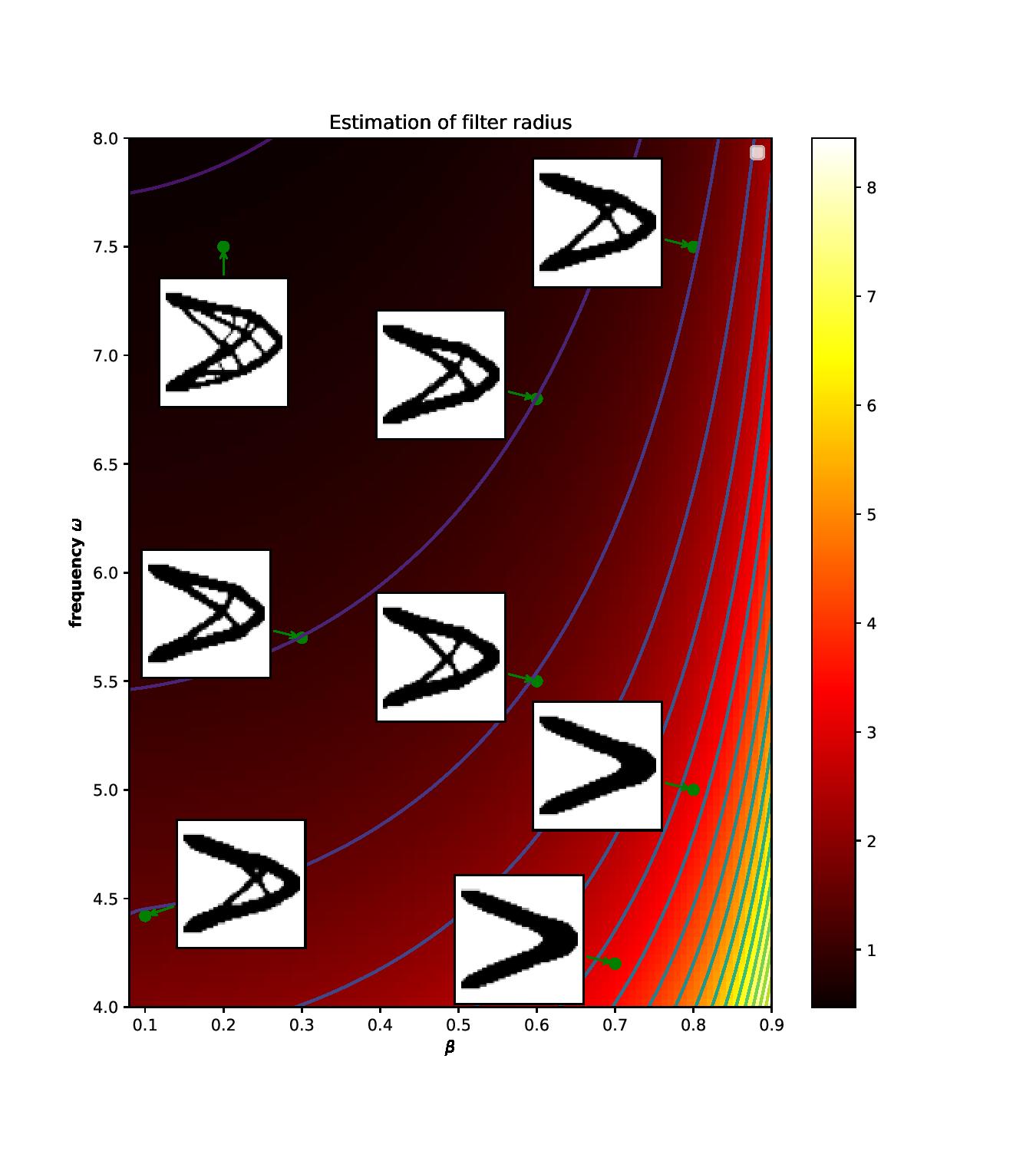}
  \caption{Colormap of $\widehat{R}_{1/2}$ in the $(\beta, \omega)$ plane, torus embedding. Level lines and shapes obtained for different radius are represented.}
  \label{filter_radius_torus}
\end{minipage}
\end{figure}

With the embeddings proposed in the last two subsections, the limiting NTK takes the form of a convolution over the input space $\mathbb{R}^d$. Figure \ref{spectral} represents the eigenvalues and eigenimages of the NTK Gram matrix $\Tilde{\Theta}_{\theta(t)}$. Even though this plot is done for a finite width network and a finite random embedding, we see that the eigenimages look like $2$D Fourier modes. The fact that the low frequencies have the largest eigenvalues supports the similarity between the NTK and a low pass filter.

This may explain why neural networks naturally avoid checkerboard patterns: the low frequencies of the shape are trained faster than the high frequencies which lead to checkerboard patterns.


\subsection{Filter radius}
\label{filter_radius}

\begin{wrapfigure}[22]{l}{0.42\textwidth}
\centering
     \includegraphics[width=0.4\textwidth, trim=1.2cm 1cm 1cm 1cm]{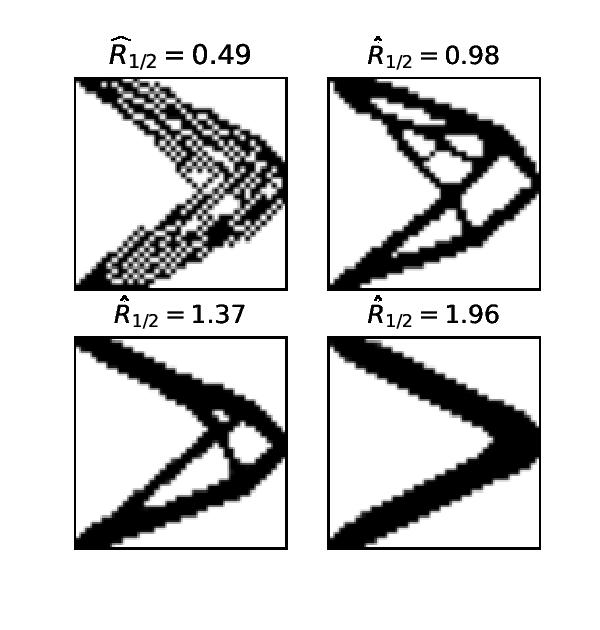}
  \captionof{figure}{Shape obtained for different values of $\widehat{R}_{1/2}$ with a Gaussian embedding for different values of $\ell \in \{ 0.5, 1, 1.4, 2 \}$.}
  \label{filter_radius}
\end{wrapfigure}


    In the classical SIMP algorithm, the choice of the radius of the filter $T$ is critical. It controls the appearance of checkerboards or intermediate densities. 
    
    When using DNNs, there is no explicit choice of filter radius, since the filter depends on  the embedding and the architecture of the network. In Section \ref{spatial_invariance} we have shown that the NTK is approximately invariant, it can hence be expressed as:
    $$
    \Theta_{\theta(t)}^L(\varphi(p),\varphi(p')) \simeq \Phi_{\infty}(\Vert p - p' \Vert),
    $$
where $\Phi_{\infty}$ can be analytically expressed with the embedding and the limiting NTK (see appendix for a detailed example).

The kernels we consider do not have compact support in general, we therefore focus instead on the radius at half-maximum of $\Phi_{\infty}$:
$$
\Phi_{\infty}(\widehat{R}_{1/2}) =  \frac{1}{2} \big( \Phi_{\infty}(0) + \inf _r \Phi_{\infty}(r) \big).
$$
Note that for simplicity we are computing here the radius of the squared filter, since obtaining a closed form formula for the square root of the NTK is more difficult. For  Gaussian filters the radius of the squared filter is $\sqrt{2}$ times that of the original, suggesting that the filter radius is well estimated by $\frac{1}{\sqrt{2}} \widehat{R}_{1/2}$.


The quantity $\widehat{R}_{1/2}$ is a function of the hyperparameters of the network ($\alpha$,$\beta$, $L$,  see appendix) and of the embedding (the lengthscale $\ell$). Using the formula for $\widehat{R}_{1/2}$, these hyperparameters can be tuned to obtain a specific filter radius. 


With the Gaussian embedding, the radius of the filter can easily be adjusted by changing the length-scale $\ell$ of the embedding. As illustrated in Figure \ref{filter_radius}.

With the torus embedding, we instead have to change the hyperparameters of the network to adjust the radius of the filter. With the ReLU activation function, the radius is very large which makes it impossible to obtain precise shape. The solution we found is to use a cosine activation $x\mapsto \cos(\omega x)$ with hyper-parameter $\omega$. Figure \ref{filter_radius_torus} shows how the radius decreases as $\omega$ increases. The $\beta$ parameter has the opposite effect, as increasing it increases the radius. For different values of $\omega$ and $\beta$, we obtain a variety of radius and plot the resulting shapes. This plot also illustrates the role of the radius in the determination of the resulting shape. The fact that cosine activation leads to an adjustable NTK radius could explain why periodic activation function help in the representation of high frequency signal as observed in \cite{periodic_activations}.

The effect of depth is more complex. For large depths $L$ the NTK either approaches a constant kernel in the so-called order regime (with infinite radius) or a Kronecker delta kernel in the so-called chaos regime (with zero radius) \cite{Chaos_Poole2016, Deep_Info_Prop_Schoenholz2017, order_chao}. Depending on whether we are in the order or chaos regime (which is determined by the activation function $\mu$ and the parameters $\alpha,\beta$), increasing the depth can either increase or decrease the radius.

We conducted an experimental study of the influence of this parameter on the geometry of the final shape. We observed that its complexity (number of holes, high frequencies) is highly controlled by $\widehat{R}_{1/2}$. We see in Figure \ref{filter_radius} and \ref{filter_radius_torus} some examples of shape obtained for several values of $\widehat{R}_{1/2}$.




\subsection{Up-sampling}

Since the density field is generated by a DNN, it can be evaluated at any point in $\mathbb{R}^d$, hence allowing upsampling. As Figure \ref{good_up_sampling} shows, with our method we obtain a smooth and binary shape. Something interesting happens when the network is trained without an embedding: when upsampling we observe some visual artifacts plotted in Figure \ref{bad_up_sampling}. We believe that it is due to the lack of spatial invariance.

Note that this second experiment was done with batch norm, as described in \cite{TOuNN}, since for this problem it was difficult to obtain a good shape with a vanilla ReLU-FCNN. With our embeddings, we can achieve complex shapes without batch-norm.



\begin{minipage}[t]{0.48\textwidth}
\centering\raisebox{\dimexpr \topskip-\height}{%
\includegraphics[scale=0.4, clip, trim = 1.5cm 2cm 1.5cm 1.5cm]{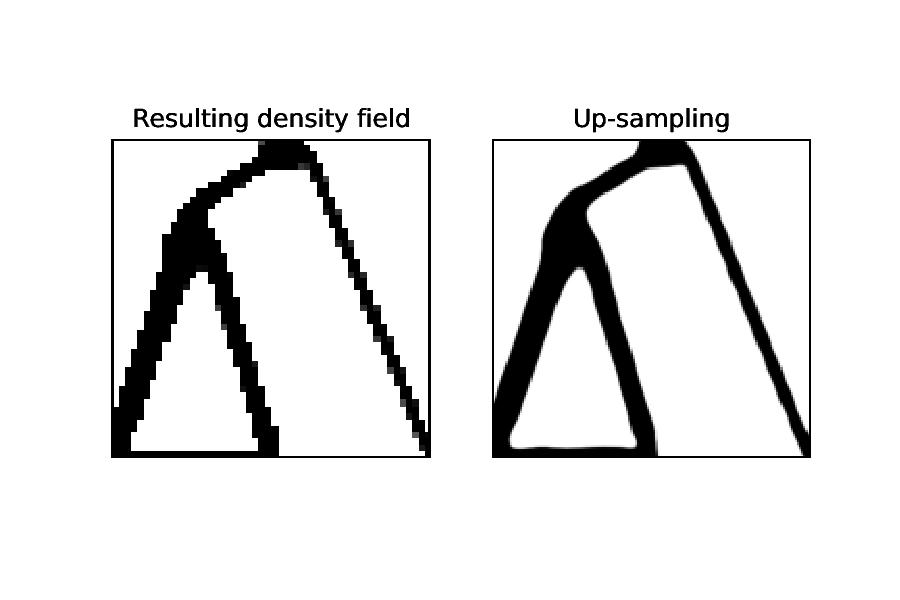}}
\captionof{figure}{Density field obtained with a Torus embedding (left) and up sampling of factor $6$ of the same network (right).}
\label{good_up_sampling}
\end{minipage}\hfill
\begin{minipage}[t]{0.48\textwidth}

 \centering\raisebox{\dimexpr \topskip-\height}{%
\includegraphics[scale=0.4, clip, trim = 0.6cm 0.9cm 1cm -0.1cm]{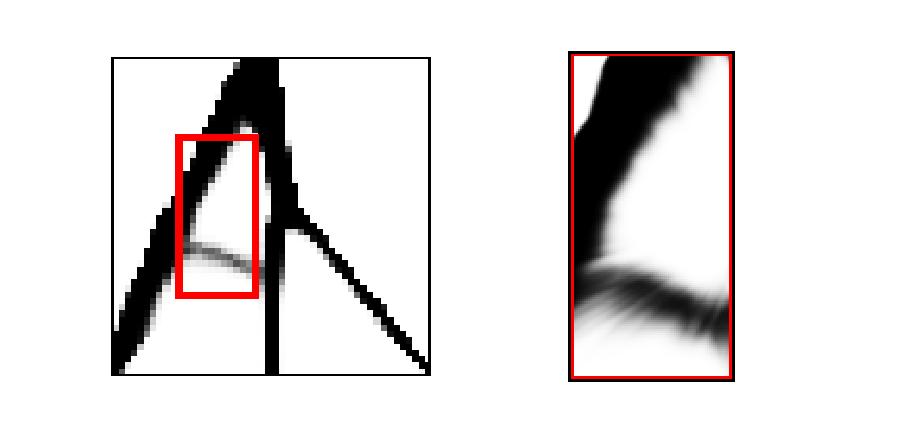}}
\captionof{figure}{Exemple of up-sampling of a FCNN (ReLu FCNN with batchnorms) without embedding, exhibing typical visual artifacts.}
\label{bad_up_sampling}
 
\end{minipage}


\section{Conclusion}
Using the NTK, we were able to give a simple theoretical description of topology optimisation with DNNs, showing a similarity to traditional filtering techniques. This theory allowed us to identify a problem: since the NTK is not translation invariant, the spatial invariance of topology optimisation is not respected, leading to visual artifacts and non-optimal shapes. We propose a simple solution to this problem: adding a spatial invariant embedding to the coordinates before the DNN.

Using this method, our models are able to learn efficient shapes while avoiding checkerboard patterns. We give tools to adjust the implicit filter size induced by the hyperparameters, to give control over the complexity of the final shape. Using the learned network, we can easily perform good quality up-sampling. The techniques described in this paper can easily be translated to any other problem where spatial invariance is needed.

The NTK is a simple yet powerful tool to analyse a practical method such as SIMP when combined with a DNN. Morover it can be used to make informed choices of the DNN's architecture and hyperparameters.

\acksection

There is no funding or competing interests associated to this work.


\bibliographystyle{plain}
\bibliography{bibliography}

\begin{thebibliography}{10}

\bibitem{topopt88}
Erik Andreassen, Anders Clausen, Mattias Schevenels, Boyan Lazarov, and Ole
  Sigmund.
\newblock Efficient topology optimization in matlab using 88 lines of code.
\newblock {\em Structural and Multidisciplinary Optimization}, 43:1--16, 11
  2011.

\bibitem{Arora}
Sanjeev Arora, Simon~S. Du, Wei Hu, Zhiyuan Li, Ruslan Salakhutdinov, and
  Ruosong Wang.
\newblock On exact computation with an infinitely wide neural net.
\newblock {\em CoRR}, abs/1904.11955, 2019.

\bibitem{CNN_topoptim}
Saurabh Banga, Harsh Gehani, Sanket Bhilare, Sagar Patel, and Levent Kara.
\newblock 3d topology optimization using convolutional neural networks.
\newblock {\em CoRR}, abs/1808.07440, 2018.

\bibitem{topology}
Bendsoe and Sigmund.
\newblock Topology optimization: Theory, methods and applications.
\newblock {\em Springer Science and Business}, April 2013.

\bibitem{Bendsoe}
Martin Bendsøe.
\newblock Bendsoe, m.p.: Optimal shape design as a material distribution
  problem. structural optimization 1, 193-202.
\newblock {\em Structural Optimization}, 1:193--202, 01 1989.

\bibitem{Fourier_TOUNN}
Aaditya Chandrasekhar and K.~Suresh.
\newblock Length scale control in topology optimization using fourier enhanced
  neural networks.
\newblock 2020.

\bibitem{TOuNN}
Aaditya Chandrasekhar and Krishnan Suresh.
\newblock Tounn: Topology optimization using neural networks.
\newblock {\em Structural and Multidisciplinary Optimization}, 2020.

\bibitem{cholmod2}
Yanqing Chen, Timothy~A. Davis, and William~W. Hager.
\newblock Algorithm 887: Cholmod, supernodal sparse cholesky factorization and
  update/downdate.
\newblock {\em ACM Transactions on Mathematical Software}, pages 1--14, 2008.

\bibitem{toward_deeper}
Amit Daniely, Roy Frostig, and Yoram Singer.
\newblock Toward deeper understanding of neural networks: The power of
  initialization and a dual view on expressivity.
\newblock {\em CoRR}, abs/1602.05897, 2016.

\bibitem{cholmod}
Timothy~A. Davis.
\newblock User guide for cholmod: a sparse cholesky factorization and
  modification package.
\newblock 2009.

\bibitem{implicit_diff}
Andreas Griewank and Christèle Faure.
\newblock Reduced functions, gradients and hessians from fixed-point iterations
  for state equations.
\newblock {\em Numerical Algorithms}, 30:113--139, 06 2002.

\bibitem{neural_param}
Stephan Hoyer, Jascha Sohl{-}Dickstein, and Sam Greydanus.
\newblock Neural reparameterization improves structural optimization.
\newblock {\em CoRR}, abs/1909.04240, 2019.

\bibitem{hierarchy}
Jiaoyang Huang and Horng{-}Tzer Yau.
\newblock Dynamics of deep neural networks and neural tangent hierarchy.
\newblock {\em CoRR}, abs/1909.08156, 2019.

\bibitem{NTK}
Arthur Jacot, Franck Gabriel, and Cl{\'{e}}ment Hongler.
\newblock Neural tangent kernel: Convergence and generalization in neural
  networks.
\newblock {\em CoRR}, abs/1806.07572, 2018.

\bibitem{order_chao}
Arthur Jacot, Franck Gabriel, and Cl{\'{e}}ment Hongler.
\newblock Order and chaos: {NTK} views on {DNN} normalization, checkerboard and
  boundary artifacts.
\newblock {\em CoRR}, abs/1907.05715, 2019.

\bibitem{metasurfaces}
Jiaqi Jiang and Jonathan~A. Fan.
\newblock Global optimization of dielectric metasurfaces using a physics-driven
  neural network.
\newblock {\em Nano Letters}, 19(8):5366–5372, Jul 2019.

\bibitem{adam}
Diederik~P. Kingma and Jimmy Ba.
\newblock Adam: A method for stochastic optimization.
\newblock 2017.

\bibitem{wide}
Jaehoon Lee, Lechao Xiao, Samuel~S Schoenholz, Yasaman Bahri, Roman Novak,
  Jascha Sohl-Dickstein, and Jeffrey Pennington.
\newblock Wide neural networks of any depth evolve as linear models under
  gradient descent.
\newblock {\em Journal of Statistical Mechanics: Theory and Experiment},
  2020(12):124002, Dec 2020.

\bibitem{topopt3D}
Kai Liu and Andres Tovar.
\newblock An efficient 3d topology optimization code written in matlab.
\newblock {\em Structural and Multidisciplinary Optimization}, 50, 12 2014.

\bibitem{heat}
Gilles Marck, Maroun Nemer, Jean-Luc Harion, Serge Russeil, and Daniel
  Bougeard.
\newblock Topology optimization using the simp method for multiobjective
  conductive problems.
\newblock {\em Numerical Heat Transfer Part B-fundamentals - NUMER HEAT
  TRANSFER PT B-FUND}, 61:439--470, 06 2012.

\bibitem{differentiable_image_param}
Alexander Mordvintsev, Nicola Pezzotti, Ludwig Schubert, and Chris Olah.
\newblock Differentiable image parameterizations.
\newblock {\em Distill}, 2018.
\newblock https://distill.pub/2018/differentiable-parameterizations.

\bibitem{topologyGAN}
Zhenguo Nie, Tong Lin, Haoliang Jiang, and Levent~Burak Kara.
\newblock Topologygan: Topology optimization using generative adversarial
  networks based on physical fields over the initial domain.
\newblock {\em CoRR}, abs/2003.04685, 2020.

\bibitem{Chaos_Poole2016}
Ben Poole, Subhaneil Lahiri, Maithra Raghu, Jascha Sohl-Dickstein, and Surya
  Ganguli.
\newblock Exponential expressivity in deep neural networks through transient
  chaos.
\newblock In D.~D. Lee, M.~Sugiyama, U.~V. Luxburg, I.~Guyon, and R.~Garnett,
  editors, {\em Advances in Neural Information Processing Systems 29}, pages
  3360--3368. Curran Associates, Inc., 2016.

\bibitem{random_features}
Ali Rahimi and Benjamin Recht.
\newblock Random features for large-scale kernel machines.
\newblock In J.~Platt, D.~Koller, Y.~Singer, and S.~Roweis, editors, {\em
  Advances in Neural Information Processing Systems}, volume~20. Curran
  Associates, Inc., 2008.

\bibitem{Rprop}
M.~Riedmiller and H.~Braun.
\newblock A direct adaptive method for faster backpropagation learning: the
  rprop algorithm.
\newblock In {\em IEEE International Conference on Neural Networks}, pages
  586--591 vol.1, 1993.

\bibitem{Rprop_full_batch}
Martin Riedmiller and Heinrich Braun.
\newblock A direct adaptive method for faster backpropagation learning: The
  rprop algorithm.
\newblock pages 586--591, 1993.

\bibitem{bochner}
W.~Rudin.
\newblock {\em Fourier Analysis on Groups}.
\newblock Wiley Classics Library. Wiley, 1990.

\bibitem{Deep_Info_Prop_Schoenholz2017}
Samuel~S. Schoenholz, Justin Gilmer, Surya Ganguli, and Jascha Sohl-Dickstein.
\newblock Deep information propagation.
\newblock 2017.

\bibitem{CGAN_topoptim}
M.{-}H.~Herman Shen and Liang Chen.
\newblock A new {CGAN} technique for constrained topology design optimization.
\newblock {\em CoRR}, abs/1901.07675, 2019.

\bibitem{filtering}
Ole Sigmund.
\newblock Morphology-based black and white filters for topology optimization.
\newblock {\em Structural and Multidisciplinary Optimization}, 33:401--424, 04
  2007.

\bibitem{periodic_activations}
Vincent Sitzmann, Julien N.~P. Martel, Alexander~W. Bergman, David~B. Lindell,
  and Gordon Wetzstein.
\newblock Implicit neural representations with periodic activation functions.
\newblock {\em CoRR}, abs/2006.09661, 2020.

\bibitem{neural_topoptim}
Ivan Sosnovik and Ivan~V. Oseledets.
\newblock Neural networks for topology optimization.
\newblock {\em CoRR}, abs/1709.09578, 2017.

\bibitem{Fourier_features}
Matthew Tancik, Pratul~P. Srinivasan, Ben Mildenhall, Sara Fridovich{-}Keil,
  Nithin Raghavan, Utkarsh Singhal, Ravi Ramamoorthi, Jonathan~T. Barron, and
  Ren Ng.
\newblock Fourier features let networks learn high frequency functions in low
  dimensional domains.
\newblock {\em CoRR}, abs/2006.10739, 2020.

\bibitem{bicgstab}
H.~Vorst.
\newblock Bi-cgstab: A fast and smoothly converging variant of bi-cg for the
  solution of nonsymmetric linear systems.
\newblock {\em SIAM J. Sci. Comput.}, 13:631--644, 1992.

\bibitem{disentangling}
Lechao Xiao, Jeffrey Pennington, and Samuel~S. Schoenholz.
\newblock Disentangling trainability and generalization in deep learning.
\newblock {\em CoRR}, abs/1912.13053, 2019.

\bibitem{optimality_criteria}
Luzhong Yin and Wei Yang.
\newblock Optimality criteria method for topology optimization under multiple
  constraints.
\newblock {\em Computers and Structures}, 79(20):1839--1850, 2001.

\end{thebibliography}

\newpage

\appendix

\section{Derivation of the algorithm}

In this section we show how to derive the equations used in our algorithm, especially the ones corresponding to implicit differentiation \cite{implicit_diff}. Let us recall that we consider a vector $X \in \mathbb{R}^N$ and compute a vector $Y = \Sigma(X) \in [0,1]^N$ (either $Y^{\text{MF}}$ or $Y^{\text{NN}}$) by:
$$
\forall i \in \{1,...,N\}, ~y_i = \sigma(x_i + \bar{b}(X)), \quad \text{such that: } \sum_{i=1}^N y_i = V_0, \quad \sigma(x) = \frac{1}{1 + e^{-x}},
$$
Where $X$ denotes $(x_1,...,x_N)$.

We want to show that this operation is well defined and find a formula to recover $\nabla_X C$ from a given $\nabla_Y C$. More precisely we have the following result.
\begin{prop}[Proposition \ref{implicit_diff} in the paper]
    \label{implicit_diff}
    Let $X \in \mathbb{R}^N$, the operation $Y = \Sigma(X)$ is well defined.
    Moreover, let $\dot{S}$ be the vector of the $\dot{\sigma}(x_i + \bar{b}(X))$. Then we have $\nabla_X C = D_X \nabla_Y C$ with:
    \begin{equation}
    \label{DX}
    D_X := -\frac{1}{\vert \dot{S} \vert_1}\dot{S}\dot{S}^T + \text{Diag}(\dot{S}).
    \end{equation}
    $D_X$ is a symmetric positive semi-definite matrix whose kernel corresponds to constant vectors and has eigenvalues smaller than $\frac{1}{2}$. 
\end{prop}
\textit{Proof: } Let us consider the function $F : \mathbb{R}^N \times \mathbb{R} \longrightarrow \mathbb{R}$ defined by: $F(z, b) = \sum_{i=1}^N \sigma(z_i + b)$.
It is clear that $F(X,.)$ is stricty increasing on $\mathbb{R}$ from $0$ to $N$. Then 
$\exists ! \bar{b} \in \mathbb{R}$ such that $F(X, \bar{b}) = V_0$.

As $\partial_b F(X, \bar{b}) > 0$, by the implicit functions theorem, there exists a neighbourhood $V$ of $X$ in $\mathbb{R}^N$, a neighbourhood $U$ of $\bar{b}$ in $\mathbb{R}$ and a function $\bar{b} : V \longrightarrow \mathbb{R}$ of class $\mathcal{C}^1$ such that:
$$
\forall (z, b) \in V \times U,~ F(z,b) = V_0 \iff b = \bar{b}(z).
$$
Moreover we also get from the implicit function theorem that: 
$$
\frac{\partial \bar{b}}{\partial x_i}(X) = - \bigg( \frac{\partial F}{\partial b}(X, \bar{b}) \bigg)^{-1} \frac{\partial F}{\partial x_i}(X, \bar{b}) = - \bigg( \sum_{j=1}^N \dot{\sigma}(x_j + \bar{b})  \bigg)^{-1} \dot{\sigma}(x_i + \bar{b}),
$$
and we can apply chain rules:
$$
\begin{aligned}
    \frac{\partial C}{\partial x_i} &= \sum_{j=1}^N \frac{\partial C}{\partial y_j}\frac{\partial y_j}{\partial x_i} \\
        &= \sum_{j=1}^N \frac{\partial C}{\partial y_j} \dot{\sigma}(x_j + \bar{b}(x)) \big( \frac{\partial \bar{b}}{\partial x_i} + \delta_{ij} \big),
\end{aligned}
$$
So that equation \ref{DX} makes sense. 
Now, if we denote $\dot{S} = (a_1,...,a_N)$, let us recall that we defined $a_i = \dot{\sigma}{x_i + \bar{b}(X)}$ where $\sigma$ is the sigmoid function. By taking any $u \in \mathbb{R}^N$, we remark that:
\begin{equation}
\label{DXu}
    \big( D_X u \big)_i = \frac{a_i}{\vert \dot{S} \vert_1} \sum_{j=1}^N a_j(u_i - u_j).
\end{equation}

We easily deduce from equation \ref{DXu} that $\ker(D_X) = \text{span}(1_N)$ and that $D_X \in S_N^+(\mathbb{R})$. Indeed:
$$
\begin{aligned}
    \forall u \in \mathbb{R}^N, \quad  u^T(D_X)u &= -\frac{1}{\vert \dot{S} \vert_1} u^T \dot{S}\dot{S}^T u^T + \sum_{i=1}^N a_i u_i^2 \\
    &=  \frac{1}{\vert \dot{S} \vert_1} \bigg\{ -\bigg( \sum_{i=1}^N a_i u_i \bigg)^2 + \bigg(\sum_{i=1}^N a_i\bigg) \bigg(\sum_{i=1}^N a_i u_i^2 \bigg) \bigg\} \\
    &=  \frac{1}{\vert \dot{S} \vert_1} \sum_{1 \leq i,j \leq N} a_i a_j u_i (u_i - u_j) \\
    &= \frac{1}{\vert \dot{S} \vert_1} \sum_{1 \leq i<j \leq N} a_i a_j (u_i - u_j)^2 \geq 0.
\end{aligned}
$$

\textbf{Eigenvalues}: We already know that $0$ is an eigenvalue with multiplicity $1$. So let $u \neq 0$ in $\mathbb{R}^N$ and $\lambda > 0$ such that: $D_X u = \lambda u$. Then we easily show:
    $$
    \forall i \in \llbracket 1,N \rrbracket,~~ \frac{a_i - \lambda}{a_i}u_i = \frac{1}{\vert \dot{S} \vert_1} \sum_{j=1}^N a_j u_j =: \langle u \rangle_a.
    $$
    If $\langle u \rangle_a = 0$, then necessarily $\lambda \in \{ a_1,...,a_N \}$
    \\
    If $\langle u \rangle_a \neq 0$, then we can assume (by normalising $u$) that $\langle u \rangle_a = 1$ and we have $u_i = \frac{a_i}{a_i - \lambda}$. Then we can replace $u_i = \frac{a_i}{a_i - \lambda}$ in the equation $\langle u \rangle_a = 1$:
    $$
     \sum_{j=1}^N a_j  =  \sum_{j=1}^N \frac{a_j^2}{a_j - \lambda}, \quad \text{which by substraction leads to} \quad F(\lambda) := \sum_{j=1}^N  \frac{a_j}{a_j - \lambda} = 0,
    $$
    By studying the function $F$, we see that $\forall \lambda > \max_i(a_i),~ F(\lambda) < 0$. Therefore an eigenvalue always satisfies the inequality:
    $$
    \lambda \leq \max \{ a_1,...,a_N\} \leq \Vert \dot{\sigma} \Vert_{\infty} = \frac{1}{4},
    $$
    The last inequality coming from the fact that $a_i = \dot{\sigma} (x_i + \bar{b}(X))$, as mentionned earlier.

\textbf{Remark:} As shown above an important property of the matrix $D_X$ is that it cancels out constants, which allows us to consider the limiting NTK up to some constant. The fact that the eigenvalues of $D_X$ are in $[0,\frac{1}{4}]$ can help to avoid exploding gradients.

\section{Equations of evolution}

We quickly show how equations $5$, $6$ and $7$ of the paper are derived. The proofs are mainly based on chain rules.

Let us first remark that the matrix $D_X$ introduced above actually corresponds to the jacobian matrix $\nabla_X \Sigma$ of the application $\Sigma : \mathbb{R}^N \longrightarrow [0,1]^N$. So we can immediately applied chain rules to $Y^{\text{NN}} = \Sigma(X(\theta))$ and get:
$$
\begin{aligned}
    \frac{\partial Y^{\text{NN}}}{\partial t} &= D_{X(\theta(t))} \frac{\partial X(\theta(t))}{\partial t} \\
    &= - D_{X(\theta(t))} \Tilde{\Theta}^L_{\theta(t)} \nabla_{X_{\theta(t)}}C \quad \text{(Gradient Descent)} \\
    &= - D_{X(\theta(t))} \Tilde{\Theta}^L_{\theta(t)} D_{X(\theta(t))}  \nabla_{Y^{\text{NN}}}C (\theta(t)) \quad \text{(By proposition \ref{implicit_diff})}.
\end{aligned}
$$

Similarly, for the MF method, we set $X = T\bar{X}$ and obtain:
$$
\begin{aligned}
    \frac{\partial Y^{\text{MF}}}{\partial t} &= D_{X(t)} \frac{\partial X(t)}{\partial t} \\
    &= D_{X(t)} T \frac{\partial \bar{X}(t)}{\partial t} \quad \text{(Linearity)} \\
    &= -D_{X(t)} T \nabla_{\bar{X}} C \quad \text{(Gradient descent)} \\
    &= -D_{X(t)} T T^T\nabla_{X} C  \quad \text{(Chain rule)}\\
    &= -D_{X(t)} T T^T D_{X(t)} \nabla_{Y^{\text{MF}}} C.
\end{aligned}
$$

\section{Details about embeddings}
\subsection{Torus embedding}
The aim of this section is to give details about properties of the limiting NTK in case of Torus embedding. As a reminder we consider the following embedding:
$$
    \mathbb{R}^2 \ni p  = (p_1, p_2) \longmapsto \varphi(p) = r(\cos(\delta p_1), \sin(\delta p_1), \cos(\delta p_2), \sin(\delta p_2));
$$
In particular we show the following proposition which basically says that $\Tilde{\Theta}_{\infty}$ is in that case a discrete convolution and derive from there its spectral properties and construct its positive semi-definite square root

\begin{prop}[Proposition \ref{torus_prop} in the paper]
     We can always extend our $n_x \times n_y$ grid and choose $\delta$ such that the embedded grid covers the whole torus (typically $\delta = \frac{\pi}{2 \max(n_x, n_y)}$ and take a $n \times n$ grid with $n = 4\max(n_x, n_y)$). Then the Gram matrix $\Tilde{\Theta}_{\infty}$ of the limiting NTK is a $2$D discrete convolution matrix. Moreover the NTK Gram matrix has a positive definite square root $\sqrt{\Tilde{\Theta}_{\infty}}$ which is also a discrete convolution matrix.
\end{prop}

\textit{proof: } We assume that we extend the grid in a $n \times n$ grid with $n \geq n_x, n_y$. Now we take $\delta = \frac{2 \pi}{n}$ and we consider the limiting NTK Gram matrix on $\varphi \big( \llbracket n, n \rrbracket \times \llbracket n, n \rrbracket \big)$. 

As $\Theta_{\infty}(\varphi(p), \varphi(p'))$ depends only on $p - p'$, we can see the limiting NTK Gram Matrix as a discrete convolution kernel $\mathcal{K}$ acting on $\mathbb{Z}/n\mathbb{Z} \times \mathbb{Z}/n\mathbb{Z}$:
$$
\Theta_{\infty} ((k,k'), (j,j')) = \mathcal{K}(k-k', j-j'),
$$
For $(k,k'), ~(j,j') \in \mathbb{Z}/n\mathbb{Z} \times \mathbb{Z}/n\mathbb{Z}$.

We see $\Tilde{\Theta}_{\infty}$ as a $n^2$ square matrix with each index in $\mathbb{Z}/n\mathbb{Z} \times \mathbb{Z}/n\mathbb{Z}$.

We introduce the Fourier vectors $\Omega_m = (e^{-i2 \pi\frac{mk}{n}})_{0 \leq k \leq n_x - 1}$. As $\Tilde{\Theta}_{\infty}$ is a $2$D convolution matrix, we classically have the following results:

The eigenvectors of $\Tilde{\Theta}_\infty$ are exactly given by:
$$
\Omega_m \boldsymbol{\otimes} \Omega_M,
$$
for $0 \leq m \leq n_x - 1$ and $0 \leq M \leq n_y - 1$, $\boldsymbol{\otimes}$ denotes the Kronecker product. The corresponding eigenvalue is given by the discrete Fourier transform $\widehat{\mathcal{K}}(m, M)$ with:
$$
\widehat{\mathcal{K}}(m,M) = \sum_{j=0}^{n - 1}\sum_{j'=0}^{n - 1} e^{-i2 \pi\frac{mj}{n}}e^{-i2 \pi\frac{Mj'}{n}} \mathcal{K}(j,j').
$$
Moreover, as the matrix $\Tilde{\Theta}_{\infty}$ is positive definite (from the positive definiteness of the NTK, \cite{NTK}) those eigenvalues verify $\widehat{\mathcal{K}}(m,M) \geq 0$ and it makes sense to write the square root of the NTK Gram Matrix as the inverse Fourier transform of the $\sqrt{\widehat{\mathcal{K}}(m,M)}$:
\begin{equation}
\sqrt{\Tilde{\Theta}_{\infty}}((k,k'), (j,j')) = \frac{1}{n^2} \sum_{m=0}^{n - 1}\sum_{M=0}^{n - 1} e^{i2 \pi\frac{m(j-k)}{n}}e^{i2 \pi\frac{M(j' - k')}{n}}  \sqrt{\widehat{\mathcal{K}}(m,M)},
\label{explicit_sqrt_torus}
\end{equation}

It is easy to see that the matrix defined by equation \ref{explicit_sqrt_torus} is symmetric and positive semi-definite. Indeed we can write $\sqrt{\Tilde{\Theta}_{\infty}}((k,k'), (j,j')) = g(k-j,k'-j')$ with $g$ the Fourier transform of a positive vector.

Moreover it follows from the (discrete) convolution theorem that $\sqrt{\Tilde{\Theta}_{\infty}}^2 = \Tilde{\Theta}_{\infty}$. Therefore $\sqrt{\Tilde{\Theta}_{\infty}}((k,k'), (j,j'))$ is indeed the positive semi-definite matrix square root of $\Tilde{\Theta}_{\infty}$.

Thus the square root of the NTK Gram matrix can be seen as a convolution filter as well (it is invariant by translation as a function of $(k-j, k'-j')$).

\subsection{Dimension of radial embeddings}
In this section we prove that feature maps associated to continuous radial kernels are either trivial or of infinite dimension. this result is what motivates discussion in section \ref{spatial_invariance} of the paper.

Let us first recall Bochner theorem (\cite{bochner}):

\begin{theorem}[Bochner]
\label{bochner}
Let $(x,y) \mapsto k(x-y)$ be a continuous shift invariant positive definite kernel on $\mathbb{R}^d$. Then it is the Fourier transform of a finite positive measure $\Lambda$ on $\mathbb{R}^d$:
$$
k(r) = \int_{\mathbb{R}^d} e^{i \omega \cdot r} d\Lambda(\omega).
$$
\end{theorem}

The function $k$ appearing in the above theorem will be called a positive definite function, according to the following definition:

\begin{defin}
    \label{positive_definite}
    Let $k : \mathbb{R}^d \longrightarrow \mathbb{R}$, then $k$ is a positive definite function when for all $n$, all $p_1,\dots , p_n \in \mathbb{R}^d$ and all $c_1,...,c_n \in \mathbb{R}$ we have:
    $$
    \sum_{1 \leq i,j \leq n}c_i c_j k(x_i - x_j) \geq 0.
    $$
\end{defin}

Moreover we will denote $SO(d)$ the set of rotations matrices of dimension $d$ and the Fourier transform (for an integrable function $\psi$):
$$
\mathcal{F}\psi (\omega) = \int_{\mathbb{R}^p} \psi(p) e^{-i \omega \cdot p} dp.
$$

Let us now recall the result that we want to prove:

\begin{prop}[Proposition \ref{embedding} in the paper]
Let $\varphi:\mathbb{R}^d \to \mathbb{R}^m$ for $d>2$ and any finite $m$. If $\varphi$ satisfies 
\begin{equation}
\label{radial_embedding}
\varphi(x)^T\varphi(x')=K( \Vert x-x' \Vert)
\end{equation}
for some continuous function $K$ then  both $\varphi$ and $K$ are constant.
We will denote $k(x-x') := K( \Vert x-x' \Vert)$.
\end{prop}

\textit{Proof:} We procede in the following way: We consider an embedding $\varphi$ as described above and we are going to show that, when $K$ is not constant, one can construct arbitrarily big linearly independent families $\varphi(p_1),\dots, \varphi(p_n)$.

For now let us take pairwise distinct $p_1,\dots, p_n \in \mathbb{R}^d$ and $c_1,\dots,c_n \in \mathbb{R}$ such that:
$$
\sum_{k=1}^n c_k \varphi(p_k) = 0.
$$
A clever choice for $p_1,\dots,p_n$ will be done later.

For any $p \in \mathbb{R}^d$ and any rotation $R \in SO(d)$ we can write:
$$
\begin{aligned}
0 = \varphi(p)^T \sum_{k=1}^n c_k \varphi(p_k) =\sum_{k=1}^n c_k K(\Vert p - p_k \Vert) &= \sum_{k=1}^n c_k K(\Vert Rp - Rp_k \Vert) \\ &= \varphi(Rp)^T \sum_{k=1}^n c_k \varphi(Rp_k).
\end{aligned}
$$
Since this is true for all $p'=Rp$ we can deduce that for all $p \in \mathbb{R}^d$ and all $R \in SO(d)$ we have:
$$
\sum_{k=1}^n c_k k(p - Rp_k) = 0.
$$
We denote by $\Lambda$ the finite measure on $\mathbb{R}^d$ given by Bochner's theorem applied on $k$.

Let us take a test function $\psi \in \mathcal{S}(\mathbb{R}^p)$ in the Schwartz space, we can write successively that for all rotation $R \in SO(d)$:
$$
\begin{aligned}
    0 &= \int_{\mathbb{R}^d} \mathcal{F}\psi(p) \sum_{k=1}^n c_k k(p - Rp_k) dp \\
        &= \int_{\mathbb{R}^d} \mathcal{F}\psi(p) \sum_{k=1}^n c_k \int_{\mathbb{R}^d} e^{i \omega \cdot (p - Rp_k)} d\Lambda(\omega) dp , \quad \text{(Bochner's theorem)}\\
        &= \int_{\mathbb{R}^d} \bigg(\sum_{k=1}^n c_k e^{-i \omega \cdot (Rp_k)} \bigg) \int_{\mathbb{R}^d} \mathcal{F}\psi(p) e^{i \omega \cdot p} dp ~ d\Lambda(\omega), \quad \text{(Fubini's theorem)}\\
        &= (2 \pi)^d \int_{\mathbb{R}^d} \psi(\omega) \sum_{k=1}^n c_k e^{-i \omega \cdot (Rp_k)} d\Lambda(\omega), \quad \text{(Fourier inversion)}
\end{aligned}
$$



As $K$ is not constant, we can find $\omega_0 \in \mathbb{R}^d \backslash \{ 0 \}$ such that for all $\epsilon > 0$ small enough we have $\Lambda \big( B(\omega_0, \epsilon) \big) > 0$ (otherwise the finite positive measure $\Lambda$ would be concentrated on $0$ and $k$ would be constant).

Let $R \in SO(d)$, if we assume that $S:=\sum_{k=1}^n c_k  e^{-i \omega_0 \cdot (Rp_k)} \neq 0$ then we can find a small enough open ball $B(\omega_0, \epsilon)$ on which $Re(S))$ and $Im(S)$ have constant sign and such that: $\vert Re(S) \vert\geq c_1 > 0$ or $\vert Im(S) \vert \geq c_1 > 0$.

We choose $\psi$ such that $\psi \geq 0$, $\psi$ has compact support in $B(\omega_0, \epsilon)$ and $\psi \geq c_2 > 0$  on $B(\omega_0, \frac{\epsilon}{2})$. 
Then we obtain a contradiction by writing $0 \geq (2 \pi)^d) c_1 c_2 \Lambda(B(\omega_0, \frac{\epsilon}{2}))$. (We separate real and imaginary parts).

This implies that:
\begin{equation}
\label{sum_exp}
\forall R \in \text{SO}(d),~ \sum_{k=1}^n c_k  e^{-i (R \omega_0) \cdot p_k} = 0,
\end{equation}
Now we take a particular choice of $(p_i)$, let $p_k = (k,0,\dots,0) \in \mathbb{R}^d$. 

Up to rotations, we can assume without loss of generality that $\omega_0 = (w,0,\dots,0)$ with $w\neq 0$. Moreover, we consider the particular case of rotations in the $2$D plane generated by $(1,0,\dots,0)$ and $(0,1,0,\dots,0)$.

Therefore, equation \ref{sum_exp} implies that:
$$
\forall \theta \in \mathbb{R},~ \sum_{k=1}^n c_k  \big(e^{-i w \cos(\theta)} \big)^k= 0,
$$
So that the polynomial $\sum_k c_k z^k$ has an infinite number of roots. Thus $c_1=\dots =c_n=0$.


\subsection{Random features embedding}
\label{embedding_dimension}
In this section we give some details about the way we define random embeddings, which is very similar but slightly different than in \cite{random_features}.

If the kernel is properly scaled (i.e. $k(0) = 1$) then $\Lambda$ defines a probability measure. That's why we introduce a probability measure $\mathbb{Q}$ and write:
$$
k(r) = k(0)\int_{\mathbb{R}^d} e^{i \omega \cdot r} d\mathbb{Q}(\omega) = k(0)\mathbb{E}_{\omega \sim \mathbb{Q}}[e^{i \omega \cdot r}].
$$
Now, following the reasoning in \cite{random_features} we consider:
$$
\varphi(p)_i = \sqrt{2 k(0)}\sin(\omega \cdot p + \frac{\pi}{4} + b)
$$
With $\omega \sim \mathbb{Q}$ and $b$ a random variable with a symmetric law (note that $\mathbb{Q}$ is also symmetric). Then we have:
$$
\begin{aligned}
    \mathbb{E}[\varphi(p)_i \varphi(p')_i] &= 2 k(0)\mathbb{E}\bigg[ \bigg( \frac{e^{i \omega.p + \frac{\pi}{4} + b} - e^{-i \omega .p - \frac{\pi}{4}} -b } {2i} \bigg) \bigg( \frac{e^{i \omega .p' + \frac{\pi}{4} + b} - e^{-i \omega .p' - \frac{\pi}{4}} -b} {2i} \bigg) \bigg] \\
    &= -\frac{k(0)}{2} \bigg( e^{i\frac{\pi}{2}} \mathbb{E}[e^{i \omega .(p + p') + 2b}] + e^{-i\frac{\pi}{2}} \mathbb{E}[e^{-i \omega. (p + p') - 2b}] \\&\quad \quad \quad \quad \quad - \mathbb{E}[e^{i \omega .(p - p')}] - \mathbb{E}[e^{-i \omega . (p - p')}]\bigg) \\
    &= k(0)\mathbb{E}[e^{i \omega. (p-p')}] \\
    &= k(p-p').
\end{aligned}
$$
Therefore we reduce the variance by drawing i.i.d. samples $\omega_1, \dots , \omega_{n_0}$ and $b_1,\dots, b_{n_0}$ as described in section $3$ and computing the mean $\frac{1}{n_0} \varphi(p)^T \varphi(p')$. By the strong law of large numbers we have the almost sure convergence:
$$
\frac{1}{n_0} \varphi(p)^T \varphi(p') \underset{n_0 \to \infty}{\longrightarrow} k(p - p'),
$$
Now we can obtain Gaussian embedding by drawing the bias from $\delta_0$ and weights from $\mathcal{N}(0, \frac{1}{\ell^2} I_d)$. from the above formulas we immediately get:
$$
k(p - p') = e^{-\frac{\Vert p - p' \Vert_2^2}{2 \ell^2}}.
$$

\section{Precise computations of the Neural Tangent Kernel}\label{appendix_precise_comp_NTK}

We now give more details about the computation of the limiting NTK and detail how we obtain the limiting kernels used in Figures $6$ and $7$ of the paper.

\subsection{Limiting NTK}

For this purpose, following several authors (\cite{NTK}, \cite{disentangling}, \cite{wide}), we need to introduce some gaussian processes and their associated kernels. For a symmetric positive kernel $\Sigma$ let us define:
\[
\left\{
    \begin{aligned}
        \mathcal{T} (\Sigma)(z,z') &= \mathbb{E}_{(X,Y)\sim \mathcal{N}(0, \Sigma_{z,z'})}\big[ \mu(X) \mu(Y)  \big] \\
        \dot{\mathcal{T}}(\Sigma)(z,z') &= \mathbb{E}_{(X,Y)\sim \mathcal{N}(0, \Sigma_{z,z'})}\big[ \dot{\mu}(X) \dot{\mu}(Y)  \big]
    \end{aligned}
\right.
\quad \text{With : } \quad \Sigma_{z,z'} = \Biggl( \begin{aligned}
                                            \Sigma(z,z) &\quad \Sigma(z,z') \\
                                            \Sigma(z,z') &\quad \Sigma(z',z')
                                        \end{aligned}  \Biggr)
                                        .
\]
Then we set $\Sigma^1(z,z') = \Theta^1_{\infty}(z,z') = \beta^2 + \frac{\alpha^2}{n_0}z^Tz'$ and we define recursively:
\begin{equation}
\label{limiting_NTK}
\sigma^{l+1} = \beta^2 + \alpha^2 \mathcal{T}(\Sigma^l), \quad \dot{\Sigma}^{l+1} = \alpha^2  \dot{\mathcal{T}}(\Sigma^l), \quad \Theta_{\infty}^{l+1} = \dot{\Sigma}^{l+1}\Theta_{\infty}^l + \Sigma^{l+1}.
\end{equation}

Using those formulas it is clear that the limiting NTK is invariant under rotation.

When neurons have constant variance, the following notion of dual activation function is often very useful:

\begin{defin}
    \label{dual}
    Let $\mu : \mathbb{R} \longrightarrow \mathbb{R}$ be a function such that $\mathbb{E}_{X \sim \mathcal{N}(0,1)}[\mu(X)^2] < +\infty$, then its dual function $\hat{\mu} : [-1,1] \longrightarrow \mathbb{R}$ is defined by:
    \[
    \hat{\mu} (\rho) = \mathbb{E}_{(X,Y) \sim \mathcal{N}(0,\Sigma_{\rho})}[\mu(X) \mu(Y)],\quad \text{With : } \Sigma_{\rho} = \Biggl( \begin{aligned}
                                                    1 &\quad \rho \\
                                                    \rho &\quad 1
                                                \end{aligned} \Biggr)
                                                .
    \]
\end{defin}

We will use some properties of the dual function, which are described in \cite{toward_deeper}.

\subsection{Another way of seeing Gaussian embedding}

As explained above (Section \ref{spatial_invariance}), the Gaussian embedding can be seen as the first hidden layer of a neural network, with the first layer untrained. Thus it actually corresponds to $\Sigma^2$ with the above notations. 

Let us consider the activation function $\mu : a \longmapsto \lambda\sin(\omega a + \frac{\pi}{4})$ and denote:
\[
\forall x,y \in \mathbb{R}^{n_0}, ~\Sigma^{1}_{x,y} = \Biggl( \begin{aligned}
                \beta^2 + \frac{1 - \beta^2}{n_0} \Vert x \Vert_2^2 &\quad \beta^2 + \frac{1 - \beta^2}{n_0} x^Ty \\
                \beta^2 + \frac{1 - \beta^2}{n_0} x^Ty &\quad \beta^2 + \frac{1 - \beta^2}{n_0} \Vert y \Vert_2^2
                          \end{aligned} \Biggr),
\]
We are looking at:
$$
\Sigma^{2}(x,y) = \beta^2 + (1 - \beta^2)\mathbb{E}_{(X,Y)\sim \mathcal{N}(0, \Sigma^{(1)}_{x,y})}[\mu(X)\mu(Y)].
$$
Let $(X,Y)\sim \mathcal{N}(0, \Sigma^{1}_{x,y})$, then $X-Y$ and $X+Y$ are normal random variables and $\mathbb{V}(X - Y) = \frac{1 - \beta^2}{n_0}\Vert x- y \Vert_2^2$. Thus, using properties of characteristic functions we get:
$$
\begin{aligned}
    \mathbb{E}[\mu(X)\mu(Y)] &= \lambda^2\mathbb{E}\bigg[ \bigg( \frac{e^{i \omega X + \frac{\pi}{4}} - e^{-i \omega X - \frac{\pi}{4}} } {2i} \bigg) \bigg( \frac{e^{i \omega Y + \frac{\pi}{4}} - e^{-i \omega Y - \frac{\pi}{4}} } {2i} \bigg) \bigg] \\
    &= -\frac{\lambda^2}{4} \bigg( e^{i\frac{\pi}{2}} \mathbb{E}[e^{i \omega (X + Y)}] + e^{-i\frac{\pi}{2}} \mathbb{E}[e^{-i \omega (X + Y)}]  - \mathbb{E}[e^{i \omega (X - Y)}] - \mathbb{E}[e^{-i \omega (X - Y)}]\bigg) \\
    &= \frac{\lambda^2}{2} \mathbb{E}[e^{i \omega (X - Y)}] \\
    &=  \frac{\lambda^2}{2} \exp \bigg\{-\frac{1}{2} \omega^2  \frac{1 - \beta^2}{n_0} \Vert x- y \Vert_2^2 \bigg\} .
\end{aligned}
$$

\subsection{Computation of the NTK used for Figure $7$ in the paper }
In this section we show how one can derived analytically the function $\Phi_{\infty}$ described in Section \ref{filter_radius}. This kind of computation can be used to derive numerically the filter radius $\hat{R}_{1/2}$ and tune the hyperparameters.

We use here a Gaussian embedding $\varphi$ of size $n_0$ with lenghtscale $\ell$ followed by one hidden linear layer (activation function $x \to \sqrt{2} \max(0,x)$) of size $n_1$ and the output layer $n_2 = 1$. We also take $\alpha^2 + \beta^2 = 1$ in those experiments, to ensure constant variance of the neurons. 

By the strong law of large numbers we have for the limiting NTK of the first layer:
$$
\Theta_{\infty}^1(\varphi(p), \varphi(p')) = \beta^2 + \frac{1 - \beta^2}{n_0} \varphi(p)^T \varphi(p') \underset{n_0 \to \infty}{\longrightarrow} \beta^2 + (1 - \beta^2) e^{-\frac{\Vert p - p' \Vert^2_2}{2 l^2}} =: G(\Vert p - p' \Vert) .
$$
For the second layer, we use the notion of dual function defined above. In the case of the standardized ReLu it is computed in \cite{toward_deeper}:
$$
\hat{r}(\rho) = \rho - \frac{\rho \arccos(\rho) - \sqrt{1 - \rho^2}}{\pi} ,\quad \rho \in [-1, 1],
$$
and:
$$
\hat{\dot{r}}(\rho) = \dot{\hat{r}}(\rho) = 1 - \frac{\arccos(\rho)}{\pi}.
$$
So that we can write, with $d = \Vert p - p' \Vert$:
$$
\Phi_{\infty}(d) = \hat{r}(G(d)) + G(d) \dot{\hat{r}}(G(d)).
$$
Therefore $\Phi_{\infty}$ only depends on $\ell$ and $\beta$. From this expression we can use standard Python libraries to approximate $\hat{R}_{1/2}$ for given values of the hyperparameters.

\subsection{Computation of the NTK used for Figure $6$ in the paper }
Now we derive an approximate of the quantity $\hat{R}_{1/2}$ used in Figure $6$ of the paper. This is a little bit more difficult than with Gaussian embedding because the rotation invariance is now only an approximation, even in the infinite-width limit.

With Torus embedding, we have $n_0=4$. The embedding is followed by two hidden linear layers with standardised cosine activation function, and then the last linear layer. We used here $r=\sqrt{2}$ $\delta = \frac{\pi}{80}$ (which is the formula suggested in the paper with $n_x = n_y = 40$). As in the case of Gaussian embedding, we set $\alpha^2 = 1 - \beta^2$. This ensures that neurons have constant variance and allows easy analytical computations.

Thanks to the Torus embedding described above, we get for the first layer:
$$
\begin{aligned}
\Theta_{\infty}^1(\varphi(p), \varphi(p')) &= \beta^2 + \frac{1 - \beta^2}{n_0}\varphi(p)^T \varphi(p') \\
&= \beta^2 + \frac{1 - \beta^2}{2}\big( \cos(\delta (p_1 - p_1')) + \cos(\delta (p_2 - p_2')) \big)
\end{aligned}
$$
As rotation invariance is not analytically correct here, we look at the limiting NTK in the direction $p_1=p_2$. which gives:
$$
\Sigma^1 (\varphi(p), \varphi(p')) = \Theta_{\infty}^1(\varphi(p), \varphi(p')) = \beta^2 + (1 - \beta^2)\cos(\delta r),
$$
with $r=\vert p_1 - p_1'\vert=\vert p_2 - p_2' \vert$.

For the next layers, we use the dual function of the standardised cosine (see \cite{toward_deeper}) given by:
$$
\hat{\mu}(\rho) = \frac{\cosh(\omega^2 \rho)}{\cosh(\omega^2)},
$$
and its derivative:
$$
\hat{\dot{\mu}}(\rho) = \omega^2 \frac{\sinh(\omega^2\rho)}{\cosh(\omega^2)},
$$

Then the limiting NTK is simply given by the following formulas:
$$
\begin{aligned}
    \Sigma^{l+1}(\varphi(p), \varphi(p')) &= \beta^2 + (1 - \beta^2) \hat{\mu}(\Sigma^{l}(\varphi(p), \varphi(p'))), \\ 
     \dot{\Sigma}^{l+1}(\varphi(p), \varphi(p')) &= (1 - \beta^2) \hat{\mu}(\dot{\Sigma}^{l}(\varphi(p), \varphi(p'))), \\ 
     \Theta_{\infty}^{l+1}(\varphi(p), \varphi(p')) &=  \Sigma^{l+1}(\varphi(p), \varphi(p')) + \dot{\Sigma}^{l+1}(\varphi(p), \varphi(p'))  \Theta_{\infty}^l(\varphi(p), \varphi(p')).
\end{aligned}
$$

This way we construct a function $\Phi_{\infty}(r)$ with $r$ an approximation of the radius and we can use it to compute numerically an approximation of $\hat{R}_{1/2}$ as before.

\section{Square root of the NTK in the case of random embedding}

\label{square_root_appendix}

We now prove that we can define a notion of a square root of the NTK. First we need a technical lemma:

\begin{Lemma}
    \label{lem}
    Let $\mu$ be a continuous function such that $\mathbb{E}_{X \sim \mathcal{N}(0,1)}[\mu(X)^2] = 1$, $C \in [0,1]$ a constant and $f\geq0$ a positive definite function (in the sense of definition \ref{positive_definite}) such that $C + f(p) \leq 1$. Then the function
    $$
    F : p \longmapsto \hat{\mu}(C + f(p)) - \hat{\mu}(C),
    $$
    is positive definite, where $\hat{\mu}$ denotes the dual function of $\mu$ (see definition \ref{dual}).
\end{Lemma}

\textit{Proof: } 

Let us take $p_1,...,p_m \in \mathbb{R}^{d}$ and $c_1,...,c_m \in \mathbb{R}$. We introduce the Hermite expansion $\sum_k a_k h_k$ of $\mu$ and write its dual function as (see \cite{toward_deeper}):
$$
\hat{\mu}(\rho) = \sum_{k=0}^{+\infty} a_k^2  \rho^k, \quad \rho \in [-1,1],
$$
Then by Bernoulli's formula:
$$
\hat{\mu}(C + f(p_i-p_j)) -  \hat{\mu}(C) = \sum_{k=1}^{+\infty} a_k^2 f(p_i - p_j) \sum_{s=0}^{k-1} C^{k-1-s} (C + f(p_i - p_j) )^s.
$$
Thus by polynomial combination with positive coefficients of positive semi-definite kernels:
$$
\sum_{i,j=1}^m c_i c_j F(p_i - p_j) = \sum_{k=1}^{+\infty} \sum_{s=0}^{k-1} a_k^2  C^{k-1-s} \sum_{i,j=1}^m c_i c_j  f(p_i - p_j)(C + f(p_i - p_j) )^s \geq 0,
$$
Which achieves the proof.


Let us recall the statement that we want to prove:
\begin{prop}[Proposition \ref{continuous_square_root} in the paper]
    Let $\varphi$ be an embedding as described in section \ref{random_embedding} of the paper, for a positive radial kernel $k\in L^1(\mathbb{R}^{d})$ with $k(0) = 1$. Then there is a filter function $g:\mathbb{R}\to\mathbb{R}$ and a constant $C$ such that for all $p,p'$:
    \begin{equation}
        \label{square_root}
    \lim_{n_0\to\infty}\Theta_{\infty}(\varphi(p),\varphi(p')) = C + (g \star g)(p-p'),
    \end{equation}
    where $\Theta_\infty$ is the limiting NTK of a network with Lipschitz, non constant, and standardized activation function $\mu$.

\end{prop}

Before writing the proof, let us make some remarks on the assumptions of this proposition and their immediate implications:
\begin{itemize}
\item We recall that the fact that $\mu$ is "standardised" means here: $\mathbb{E}_{X \sim \mathcal{N}(0,1)}[\mu(X)^2] = 1$. 

\item As mentioned before (Section 
\ref{SIMP_with_DNNs} of the paper) we assume for simplicity that $\alpha^2 = 1 - \beta^2$ to ensure constant variance of the neurons (we consider  $\beta \in [0,1)$).

\item We denote by $A$ the Lipschitz constant of $\mu$. By Rademacher theorem, we know that $\mu$ is almost everywhere differentiable and $\Vert \dot{\mu} \Vert_{\infty} \leq A$. The fact that $\mu$ is not constant ensures that $\hat{\mu}$ is (strictly) increasing on $[0,1)$.

\item Moreover, the Lipschitz assumption also implies that $\vert\hat{\dot{\mu}}(1)\vert \leq A^2 < +\infty$ and therefore $\hat{\dot{\mu}}$ is continuous on $[-1,1]$ by Abel's theorem on entire series.

\item The procedure to approximate the kernel $k$ in Section \ref{random_embedding} of the paper assumes that $k$ is continuous (to be able to apply Bochner's theorem). It is therefore also the case in this proof. 

\end{itemize}

\textit{Proof of the proposition:}

\textbf{Step $1$:} We want to show by recursion that for all $l \geq 1$ there exists some constant $C_l \in [0, 1)$ such that for all $p,p' \in \mathbb{R}^{d}$ we have in probability:
\begin{equation}
    \label{rec_sigma}
  \Sigma^{l}(\varphi(p),\varphi(p')) \underset{n_0 \to \infty}{\longrightarrow} C_l + f_l(p-p'),
\end{equation}
With $f_l$ a radial positive definite function such that $f_l \geq 0$ and $f_l \in L^1(\mathbb{R}^{d})$.

For $l = 1$, we know that this is true by the law of large numbers:
\begin{equation}
\label{recurrence_initialisation}
\begin{aligned}
\Sigma^1(\varphi(p),\varphi(p')) = \Theta_{\infty}^1(\varphi(p),\varphi(p')) &= \beta^2 + \frac{1 - \beta^2}{n_0} \varphi(p)^T \varphi(p') \\
&\underset{n_0 \to \infty}{\longrightarrow} \beta^2 + (1 - \beta^2)k(p-p'),
\end{aligned}
\end{equation}
We just set $f_1 = (1 - \beta^2)k$. Now we assume $l \geq 2$:

We have by our normalisation assumptions $\Sigma^{l}(\varphi(p),\varphi(p)) = C_l + f_l(0) = 1$. Using the continuity of $\hat{\mu}$ (see \cite{toward_deeper} for the properties of $\hat{\mu}$), we have:

\begin{equation}
\label{convergence_step_1}
\begin{aligned}
\Sigma^{l+1}(\varphi(p),\varphi(p')) &= \beta^2 + (1 - \beta^2) \hat{\mu}(\Sigma^{l}(\varphi(p),\varphi(p'))) \\
& \underset{n_0 \to \infty}{\longrightarrow} \beta^2 + (1 - \beta^2) \hat{\mu} (C_l + f_l(p-p')).
\end{aligned}
\end{equation}

Using properties of the dual function given in \cite{toward_deeper}, we know that $\hat{\mu}$ is positive, increasing and convex in $[0,1]$. Moreover as $f_l$ is radial positive definite we have $f_l \leq f_l(0) = 1 - C_l$. Then by convexity:
$$
\begin{aligned}
    \hat{\mu}(C_l + f_l(p-p')) &= \hat{\mu}\bigg(\frac{f_l(p-p')}{1 - C_l}  + \bigg( 1 - \frac{f_l(p-p')}{1 - C_l} \bigg)C_l\bigg)\\
    &\leq \frac{f_l(p-p')}{1 - C_l}\hat{\mu}(1) + \bigg( 1 - \frac{f_l(p-p')}{1 - C_l} \bigg)\hat{\mu}(C_l).
\end{aligned}
$$
Using that $\hat{\mu}$ is increasing:
$$
\vert  \hat{\mu}(C_l + f_l(p-p')) -  \hat{\mu}(C_l) \vert \leq \frac{\hat{\mu}(1) - \hat{\mu}(C_l)}{1 - C_l}f_l(p-p'),
$$
So that we can rewrite equation \ref{convergence_step_1} in the following form:
$$
\Sigma^{l+1}(\varphi(p),\varphi(p')) \underset{n_0 \to \infty}{\longrightarrow} \beta^2 + (1 - \beta^2) \hat{\mu}(C_l) + f_{l+1}(p-p'),
$$
With $f_{l+1}(p-p') = (1 - \beta^2) (   \hat{\mu}(C_l + f_l(p-p')) -  \hat{\mu}(C_l)  )$ and $C_{l+1} =  \beta^2 + (1 - \beta^2) \hat{\mu}(C_l)$.

The previous inequality, lemma \ref{lem} and the fact that $\hat{\mu}$ is increasing in $[0,1)$ ensure the properties of $f_{l+1}$ and $C_{l+1}$.

\textbf{Step $2$:} As $\hat{\dot{\mu}}$ is also positive, continuous, increasing and convex in $[0,1]$, we can obtain a convergence in probability similar to equation \ref{rec_sigma} but for $\dot{\Sigma}^{l}$:
$$
\dot{\Sigma}^{l}(\varphi(p),\varphi(p')) \underset{n_0 \to \infty}{\longrightarrow} B_l + h_l(p-p'),
$$
With $B_l \geq 0$, and $h_l$ a positive definite function such that $h_l \in  L^1(\mathbb{R}^{d})$ and $h_l\geq 0$.

Now we want to show by recursion that for a fixed $l$:
\begin{equation}
    \label{rec_theta}
     \forall p,p' \in \mathbb{R}^{d},~ \Theta_{\infty}^{l}(\varphi(p),\varphi(p')) \underset{n_0 \to \infty}{\longrightarrow} C_{\mu, \beta, l} + \theta_l(p-p').
\end{equation}
With $\theta_l$ a positive definite function such that $\theta_l \in L^1(\mathbb{R}^{d})$ and $C_{\mu, \beta, l} \geq 0$. Again we know that this is true for $l=1$ by equation \ref{recurrence_initialisation}.

We have:
$$
\Theta_{\infty}^{l+1}(\varphi(p), \varphi(p')) \underset{n_0 \to \infty}{\longrightarrow} (C_{\mu, \beta, l} + \theta_l(p-p')) \dot{\Sigma}^{(l+1)}(p,p') + C_{l+1} + f_{l+1}(p-p').
$$
So that we can set:
$$
\theta_{l+1}(\varphi(p),\varphi(p')) =  C_{\mu, \beta, l} h_{l+1}(p-p') + \theta_l(p-p')\dot{\Sigma}^{l+1}(\varphi(p), \varphi(p')) + f_{l+1}(p-p'),
$$
and:
$$
C_{\beta, \mu, l+1} = C_{l+1} + C_{\beta, \mu, l}B_l.
$$
Using that $\vert\theta_l(p-p')\dot{\Sigma}^{l+1}(\varphi(p), \varphi(p'))\vert \leq A^2 \vert \theta_l(p-p') \vert$ and all the previous results, the recursion works automatically and we have equation \ref{rec_theta} for all $l\geq 2$.

Moreover $(p,p') \longmapsto \theta_l(p-p')\dot{\Sigma}^{1 + l}(p,p') $ is positive semi-definite as a product of two positive semi-definite kernels. By sum we deduce that $\theta_{l+1}$ is positive semi-definite and by recursion we have the result for all $\theta_l$.

\textbf{Step $3$:} Now, using integrability of $\theta_l$, we know that its Fourier transform defines a function $q \in L^{\infty}(\mathbb{R}^d)$. 

From dominated convergence theorem we deduce that $q$ is continuous.

Therefore in the sense of distributions, the Fourier transform of $\theta_L$ is given by a finite positive measure (Bochner's theorem) and also by $q \in L^{\infty}(\mathbb{R}^d)$. We deduce that $q$ is the density of this finite positive measure (the Radon-Nikodym derivative with respect to the Lebesgue measure). 

From those arguments we get $q \ge 0$ and $q \in L^1(\mathbb{R}^d)$. We then have the Fourier inversion formula for $\theta_L$:

$$
\theta_L(p-p') = \frac{1}{(2 \pi)^d} \int_{\mathbb{R}^{d}} q(\omega) e^{i \omega . (p-p')} d\omega, \quad \text{with: } q\geq 0
$$
Hence it makes sense to define:
$$
g = \mathcal{F}^{-1}(\sqrt{q}),
$$
In the sense of the Fourier transform of a $L^2$ function. Then the convolution theorem ensures: 
$$
\theta_L = g \star g.
$$

\textbf{Remark:} Here we used lemma \ref{lem} and the dual activation function to show that both $f_l$ and $\theta_l$ are positive definite. If we only show that $\theta_l \in L^1(\mathbb{R}^d)$ it is still possible to show the same properties of the function $q$ by using positive definiteness of $C + \theta_L$ and take the Fourier transform in the sense of distributions, which leads to $(2\pi)^d C \delta_0 + q = (2 \pi)^d M$ with $M$ a finite positive measure. Then arguments based on test functions and the continuity of $q$ give the result. The advantage of lemma \ref{lem} is that it is a bit more general.

\section{Additional experimental results}
\label{additional_results}



\subsection{Plots of the Neural Tangent Kernel} 
Here are some additional experimental results regarding the comparison between the theoretical (limiting) NTK $\Tilde{\Theta}_{\infty}$ and the empirical NTK $\Tilde{\Theta}_{\theta(t)}$. Here again the "lines" of the Gram matrices are reshaped as images.

\begin{figure}[!h]
    \centering
    \includegraphics[scale=0.5, clip, trim=2cm 1cm 1cm 0.5cm]{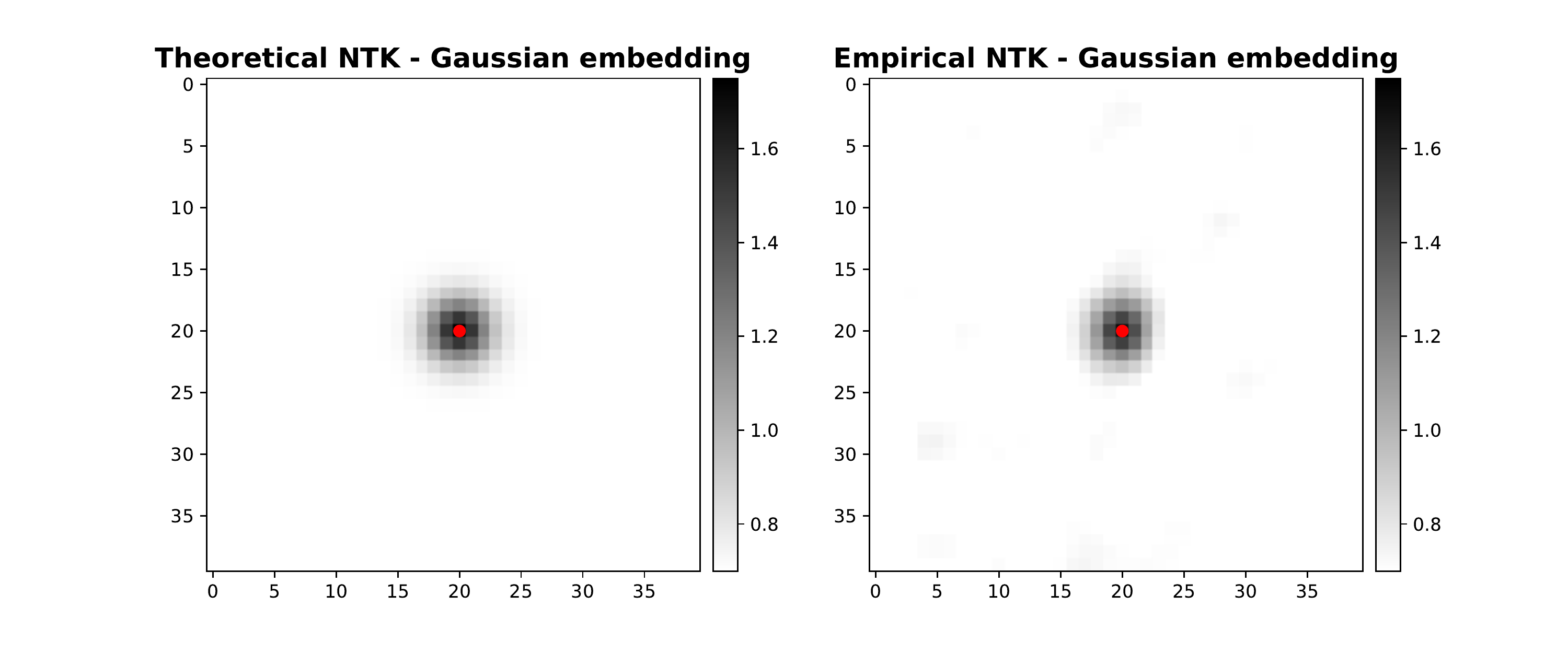}
    \caption{Comparison between one line of the Gram matrix of the empirical NTK $\Tilde{\Theta}_{\theta(t)}$ and and of the corresponding limiting NTK $\Tilde{\Theta}_{\infty}$. Here we use a Gaussian embedding as described in the paper}
    \label{ntk_comparison}
\end{figure}

Figure \ref{ntk_comparison} represents the comparison between the limiting NTK and the emprirical NTK with a Gaussian embedding. We can observe that the infinite-width limit seems to be well-respected.
 
\begin{figure}[!h]
    \centering
    \includegraphics[clip, trim=5cm 0cm 4cm 1cm ,scale=0.4]{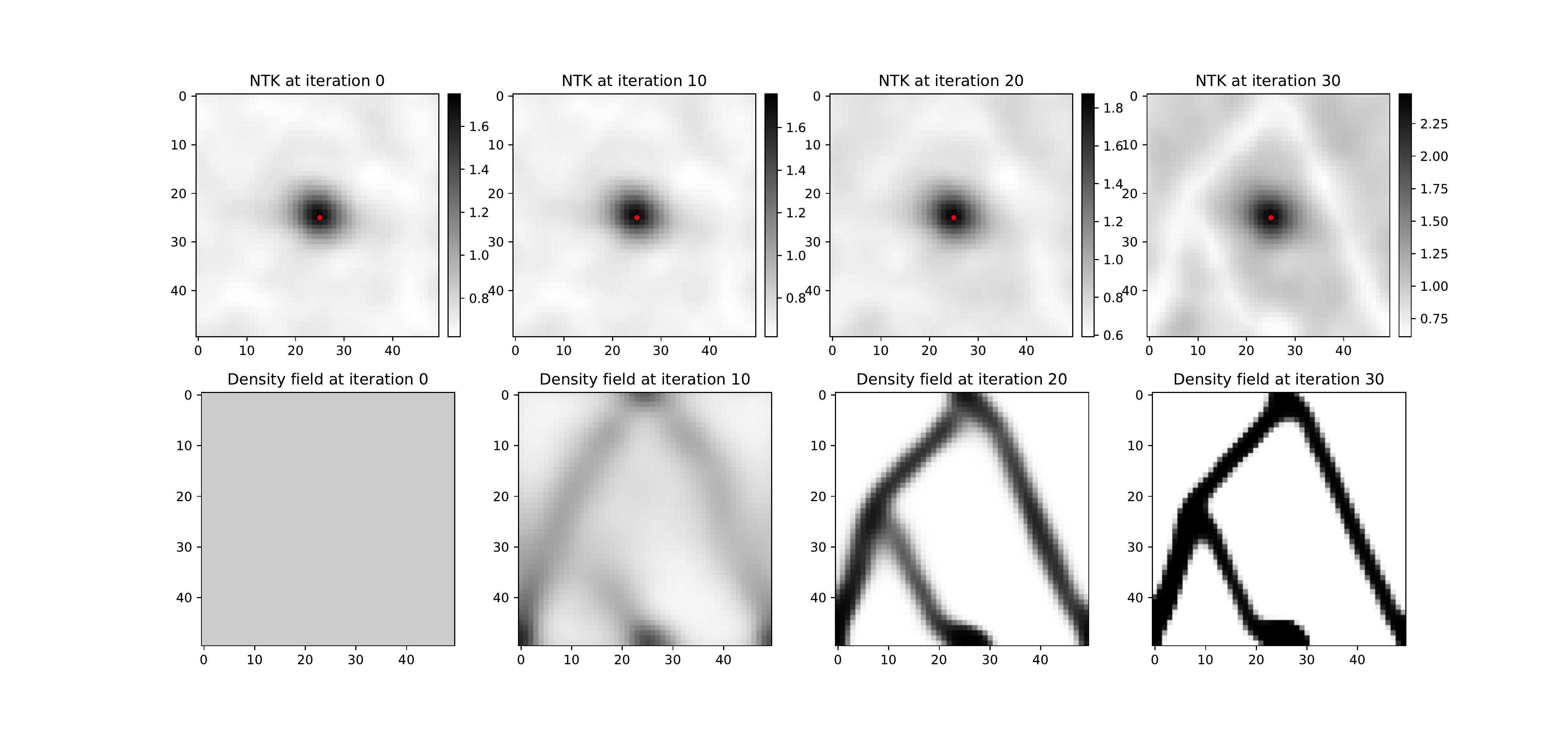}
    \caption{Evolution of the NTK of a network with a Gaussian embedding with hyperparameters as described in Section \ref{experimental_setup}. We can see a relative stability of the NTK}
    \label{ntk_evolution_img}
\end{figure}

Figure \ref{ntk_evolution_img} shows the evolution of the NTK during the optimisation process. While the NTK begins to change at the end of training (it is due to the alignment of descent directions, because of the sigmoid we use to control the volume, pre-densities $(x_i)_{1\leq i \leq N}$ tend to infinity) the NTK stays close to $\Theta_{\infty}$ during the part of training  where the final shape is created. This justifies even more that it is pertinent to study the effect of the NTK on the final geometry.

\end{document}